\documentclass[sigconf]{acmart}
\usepackage{subcaption}
\AtBeginDocument{%
  }

\setcopyright{acmlicensed}
\copyrightyear{.}
\acmYear{.}
\acmDOI{}

\begin{document}

\title[Too Big to Inspect? Ethically Auditing Large Datasets Through Small Sample]{Too Big to Inspect?\\Ethically Auditing Large Datasets Through Small Samples}

\author{Giovana Marchetti$^{1}$, Jhessica Silva$^{1}$, Tiago Torrent$^{2}$, Carlos Caetano$^{1}$, Sandra Avila$^{1}$\vspace{0.15cm}}

\authornote{Corresponding author: S.~Avila: sandra@ic.unicamp.br.}


\affiliation{
            \institution{$^{1}$Instituto de Computação (IC), Universidade Estadual de Campinas (UNICAMP), Brazil}}

\affiliation{
            \institution{$^{4}$Faculdade de Letras, Universidade Federal de Juiz de Fora (UFJF), Brazil}}

\renewcommand{\shortauthors}{Marchetti et al.}

\begin{abstract}  
  Big techs and researchers increasingly claim that their training datasets are too large to inspect for bias and ethical problems. 
  However, in the race to build ever-large models, they have relied on automated web scraping with little or no human curation. Datasets built this way absorb the stereotypes, exclusionary norms, and historical prejudices of their sources, and the models trained on them can reproduce and amplify those patterns. 
  Because manually auditing billions of samples is humanly impossible, we need practical, effective methods to identify biases in these datasets. 
  This paper studies the feasibility of ethically auditing large datasets through small samples and proposes a practical method for identifying structural biases by applying AI Ethics Tools (AIETs) to random samples. We applied Datasheets for Datasets, Data Statements, Data Nutrition Labels, and Data Cards to a 1\% sample of Framed Multi30K, and Datasheets for Datasets to two 0.00006\% samples of relaion2B-en-research-safe. 
  Our results show that evaluating small samples with AIETs can reveal ethical problems, structural biases, and technical inconsistencies.  
  Although industry practice assumes that identifying problems requires analyzing the entire dataset, we show that a random subset supports a viable qualitative analysis without overwhelming computational or human resources. Demonstrating that a dataset is biased does not require examining every instance: if a bias appears in a random sample, it is a flaw of the creation process, not an anomaly.  
  The AIETs’ questions acted as ``provocateurs'', evoking different perspectives and analysis methods for the same dataset, but the analysis and its results depended fundamentally on the human factor. 
  We hope these findings contribute to urgent public discussions about ethics, transparency, accountability, and AI regulation by offering a practical pathway for responsibly curating and auditing massive datasets.
\end{abstract}

\keywords{Audit, Dataset, Dataset documentation}

\maketitle

\section{Introduction}

``Too big to inspect'' is a dodge. Driven by popular scaling laws, the pursuit of ever-larger datasets has become widespread across industry and academia, often under the assumption that more data inherently leads to better models~\cite{kaplan2020scaling, hoffmann2022training}. However, uncurated dataset expansion actively amplifies severe risks, embedding toxic content and systemic bias at scale~\cite{bender2021dangers, birhane2023hate,gebru2021datasheets,birhane2024dark}. This race increases the creation of massive datasets, usually collected automatically from the internet or annotated via crowd-sourcing, without proper curation~\cite{dodge2021documenting, luccioni2021whats, paullada2021data}, consent~\cite{longpre2024consent}, or respect for copyright~\cite{longpre2024large, bandy2021addressing}. Dataset creators and developers justify their lack of curation by citing the size of the dataset, arguing that with so much data, they cannot audit it, manually identify bias, or process it without massive computational resources, creating a severe documentation debt~\cite{bender2021dangers}. This is a flawed, convenient excuse to shift responsibility downstream~\cite{sambasivan2021everyone}. Just as statistics allows us to predict an election without interviewing every voter, or test a polluted river without drinking it whole, we do not need to analyze every byte of a dataset to audit it. We only require that those who create, apply, and deploy these datasets understand their potential content and failure modes, a necessity to set appropriate usage boundaries, drive system iterations, and alert stakeholders of \mbox{potential harm~\cite{gebru2021datasheets, raji2021ai}.}
 
These uncurated datasets directly power AI systems deployed across societal domains. For example, in employment, Amazon was forced to abandon an automated AI recruiting tool after male-dominated training data led the system to systematically penalize female applicants~\cite{dastin2018amazon}. In healthcare, a widely used commercial algorithm prioritized healthier white patients over sicker Black patients because it relied on healthcare costs as a proxy for medical need, failing to account for systemic disparities in access to care~\cite{obermeyer2019dissecting}. Similar representational gaps cause dermatological AI to perform poorly on darker skin tones~\cite{wen2022characteristics}, with state-of-the-art models suffering performance drop-offs even with a diverse dataset created exactly to feature diverse skin tones~\cite{daneshjou2021disparities}. Beyond predictive models, generative AI trained on uncurated web scraping actively reproduces racist and misogynistic stereotypes, associating certain professions (e.g., a ``CEO'' or ``judge'') with white men and criminalizing dark-skinned individuals~\cite{nicoletti2023humans,luccioni2023stable}. At its extreme, unexamined scaling poses severe safety hazards: LAION-5B~\cite{schuhmann2022laion}, a massive dataset widely used in the AI field, was temporarily taken down after researchers discovered thousands of non-consensual explicit images and child sexual abuse images~\cite{birhane2021multimodal}. This shows that treating web data as ``neutral ground truth'' obscures these real-world harms.
  
In this context, AI Ethics Tools (AIETs) provide practical frameworks to identify potential harms during AI design, development, and deployment, offering actionable pathways for mitigation~\cite{silva2026evaluation}. By applying ethical values throughout the AI lifecycle, AIETs help developers, organizations, companies, governments, regulators, and other stakeholders build trust, transparency, and accountability. Among these are data-centric documentation tools, such as \textit{Datasheets for Datasets}~\cite{gebru2021datasheets}, \textit{Data Statements}~\cite{bender2018data,10.1145/3594737}, \textit{Dataset Nutrition Label}~\cite{chmielinski2022dataset, holland2018}, and~\textit{Data Cards} \cite{pushkarna2022data}, which assist in the understanding and processing of data used in model training and evaluation.

Despite the promise of AIETs, operationalizing ethical data documentation remains a major challenge for dataset builders. In an industry that prioritizes rapid deployment, developers often see data documentation as an overhead. When confronted with web-scale, multimodal datasets comprising billions of samples, applying a tool like~\textit{Datasheets} can feel intractable, a barrier compounded by a lack of standardized ethical documentation norms~\cite{gebru2021datasheets} and limited practitioner reflection regarding societal impacts~\cite{BPLN_livro_4ed:2026}. In this paper, we argue that inspecting every individual instance in a dataset is unnecessary to identify biases in these datasets. Instead, applying AIETs to small, random samples enables effective risk identification without overwhelming computational or human resources. Crucially, if bias appears in a random sample, it reflects a structural flaw in the creation process rather than an isolated anomaly.

To validate this approach, we applied four distinct AIETs (\textit{Data\-sheets for Datasets}~\cite{gebru2021datasheets}, \textit{Data Statements}~\cite{bender2018data,10.1145/3594737}, \textit{Dataset Nutrition Label}~\cite{chmielinski2022dataset,holland2018}, and \textit{Data Cards}~\cite{pushkarna2022data}) to a 1\% sample of the multimodal and multilingual \textit{Framed Multi30K} dataset~\cite{viridiano2024framed}. Additionally, we applied the \textit{Datasheets for Datasets} tool to fewer than 1,200 samples representing 0.00006\% of the massive \textit{relaion2B-en-research-safe} dataset~\cite{relaion}. Our results demonstrate that even a small sample can reveal structural biases, ethical risks, and curation flaws in massive datasets. 
While analyzing a sample cannot uncover every isolated bias across an entire dataset, this approach enables developers to systematically identify critical ethical risks and representational harms prior to release. Practitioners can then document known failure modes, define appropriate usage boundaries, and warn users of potential risks. 
We hope this work provides a practical, scalable pathway for dataset auditing, contributing to ongoing discussions on AI governance, transparency, and responsible data curation.

In summary, our contributions are:

\begin{enumerate}
    \item We introduce a methodology for identifying bias and ethical problems in massive datasets. To do this, we apply AIETs to small, random samples, showing that ethical auditing is possible for any dataset, \mbox{regardless of~size.}
    \item We apply this methodology to audit a structured, annotated, and curated dataset (\textit{Framed Multi30K}~\cite{viridiano2024framed}), revealing that translation, annotation, and curation create ethical problems, such as representational harm and gender bias, demonstrating that human curation does not resolve all bias.
    \item We further audit a massive, web-scale, uncurated and filter dataset (\textit{relaion2B-en-research-safe}~\cite{relaion}), revealing that even in a tiny sample, ethical problems persist, such as inefficient automated safety filters, exposed personal information, and societal stereotypes, indicating that ethical auditing is viable for every dataset.
\end{enumerate}

\section{Background and Related Work}
\label{sec:back_rwork}
In this section, we overview the data-centric AIETs evaluated in this study alongside closely related work. In Section~\ref{subsec:aiets}, we introduce the selected AIETs, their creation, content, and intended use. In  Section~\ref{subsec:relatedw}, we then explore the literature on massive dataset auditing, contrasting previous automated and large-scale analyses with our proposed pipeline, which relies on manual analysis and small samples.

\subsection{AI Ethics Tools}
\label{subsec:aiets}

We identified and evaluated AIETs for multimodal text and image datasets by reviewing the literature and screening the survey by~\citet{reynolds2026reflection}. We applied the following filtering steps: (1)~we removed studies on documentation tools, (2)~we screened the remaining entries by title, (3)~we removed automated tools, (4)~we screened by abstract and quick read, and (5)~we read the remaining works in full. 
This process yielded four tools: \textit{Datasheets for Datasets}~\cite{gebru2021datasheets}, \textit{Data Statements}~\cite{10.1145/3594737}, \textit{Data Nutrition Label}~\cite{chmielinski2022dataset}, and \textit{Data Cards}~\cite{pushkarna2022data}. We selected them because each is comprehensive enough to apply across diverse datasets, has become a reference in the AI transparency literature, and takes a distinct approach to data documentation.\vspace{-0.1cm}

\paragraph{\textbf{Datasheets for Datasets}}

Proposed by~\citet{gebru2021datasheets}, \textit{Data\-sheets for Datasets} is one of the most influential and widely used dataset documentation tools in data transparency. The tool draws on the datasheets of the electronics industry, which describe the technical  specifications, behavior, limitations of hardware components. It aims to provide every dataset with standardized lifecycle documentation, helping dataset creators and users communicate. 
Although it targets general-purpose datasets and does not focus on a specific modality, its structure inspired and served as a basis for several documentation methodologies in specific AI domains.\vspace{-0.1cm}

\paragraph{\textbf{Dataset Nutrition Label}}

\citet{holland2018} originally developed the \textit{Dataset Nutrition Label}, and \citet{chmielinski2022dataset} proposed its second version, which translates statistical complexity into a accessible interface and centers on user experience. The tool draws on nutritional information labels on food and provides data users with a quick and visual way to evaluate the `` ingredients" in a data set before they use it. Like \textit{Datasheets for Datasets}, it targets general-purpose use and applies to various types of datasets. It stands out because it turns the technical questionnaire into a visual representation that lets users quickly consult critical facts, judge whether  a dataset suits their needs, and avoid inappropriate \mbox{use in~models.}\vspace{-0.1cm}

\paragraph{\textbf{Data Statements}}

\citet{bender2018data} and \citet{10.1145/3594737} proposed \textit{Data Statements}, now in its third version, to document linguistic resources for Natural Language Processing (NLP). 
Unlike technical and objective questionnaires, it asks reflective, qualitative, and discursive questions about the populations behind the data, including speakers and annotators, to reduce the exclusion of linguistic varieties and demographic groups. A 42-page guide is also provided, describing best practices and specific guidelines on how to complete this tool.\vspace{-0.1cm}

\paragraph{\textbf{Data Cards}}
Proposed by \citet{pushkarna2022data}, this tool is a framework that documents datasets for practical industrial and research use. It gives stakeholders structured summaries of essential facts they need across a dataset's lifecycle to foster responsible AI development. The tool captures both observable dataset attributes and unobservable human processes, explanations, and rationales that shape the data. 
Although it adapts to various data modalities, it centers on an accessible artifact that supports decisions about accountability, utility, quality, risks, and the consequences of using the dataset.\vspace{-0.1cm}

\subsection{Related Work}
\label{subsec:relatedw}

Extensive research focuses on auditing massive datasets to expose problems such as representation bias, toxicity, copyright issues, and security flaws.

Audits of text corpora show that neither scale nor curation guarantees quality. \citet{dodge2021documenting} analyzed the entire Colossal Clean Crawled Corpus (C4)~\cite{raffaelc4} with matching scripts, domain clustering, rule-based filters, and manual inspection of specific cases, and found benchmark contamination (over 50\% of test examples from benchmarks like GLUE~\cite{wang2018glue} in the training data), machine-translated text, and patent dumps. \citet{luccioni2021whats} audited a 1\% random sample of Common Crawl with hate speech classifiers, n-gram approaches, and perplexity scoring, finding that more than 2\%~of pages contained sexually explicit material and 4\% to 5\% contained hate speech. Even a curated dataset is not free of problems: \citet{bandy2021addressing} documented the entire BookCorpus~\cite{zhu2015bookcorpus} using \textit{Datasheets for Datasets}~\cite{gebru2021datasheets}, and, with string-matching, hashing, and manual checks, found that 35\% of the books were duplicates, that genres were heavily skewed, and that some books carried explicit copyright prohibition notices.

Multimodal audits, mostly of the LAION family, report similar problems with added visual and social harms. \mbox{\citet{birhane2021multimodal}} used CLIP-retrieval and k-nearest neighbors (k-NN) searches on LAION-400M~\cite{schuhmann2021laion} and showed that some demographic terms (``la\-tina'', ``black woman''), relationship identifiers (``aunty'', ``mommy''), and professions (``CEO'') consistently returned pornography, non-con\-sensual images, sexual violence, and Eurocentric stereotypes. \citet{birhane2023into} later compared LAION-400M and LAION-2B-en~\cite{schuhmann2022laion} with an automated hate speech classifier and found that hate speech increased by 12.26\% as the dataset grew. \citet{dominguez2026unmasking} applied a computer vision pipeline on 79,902 faces extracted from about 464,000 LAION-5B images, finding a dataset dominated by young white males, with gendered emotional stereotypes such as ``anger'' associated mostly with men and ``happiness'' with women. Beyond bias, \citet{webster2023duplication} found about 700 million duplicated images in LAION-2B using nearest neighbor search, which helps explain why models like Stable Diffusion memorize and reproduce training images~\cite{somepalli2023diffusion}. At the legal extreme, \citet{thiel2023identifying} combined perceptual hashing (PhotoDNA), MD5 matching against law enforcement datasets, k-NN embedding analysis, and manual verification to find over 3,000~instances of child sexual abuse material in LAION-5B~\cite{schuhmann2022laion}. This forced LAION to take the dataset down in December 2023 and release \textit{Re-LAION-5B-research}~\cite{relaion}.

Filtering is the standard response to these problems, but audits show it creates new ones. In C4, the blocklist meant to remove ``bad'' words also erased content from marginalized groups, LGBTQ+ topics, and dialects like African-American English~\cite{dodge2021documenting}. Perplexity filtering seems to remove sexual content from Common Crawl but not hate speech~\cite{luccioni2021whats}, and image-based safety filters (\textit{p\_unsafe}) miss problematic text~\cite{birhane2023into}. \citet{hong2024s} describe an ``exclusion amplification'' effect in CLIP-filtering on DataComp CommonPool: the filter disproportionately discards images of marginalized groups (like LGBTQ+ people) and of non-Western regions, yet keeps copyrighted stock photos and millions of sexually explicit images. These dataset problems also carry over into trained models: \citet{birhane2024dark} found that OpenCLIP models trained on larger LAION datasets are much more likely to misclassify Black and Latino faces as~\mbox{``criminals''.}

Despite their contributions, these audits share four limitations. First, they are computationally expensive: they downloaded terabytes of data, used large GPU clusters, and ran millions of instances through automated pipelines~\cite{birhane2023into, dominguez2026unmasking, webster2023duplication}. We show this is not necessary: a small random sample (1,200 instances, or 0.00006\% of relaion2B-en-research-safe; 316 instances, or 1\% of Framed Multi\-30K) evaluated on standard hardware is enough to find bias and \textit{(un)}curation problems. Second, many rely on targeted searches, keyword lists, or hash matching~\cite{birhane2021multimodal, thiel2023identifying}, which requires knowing the bad words or illegal files in advance; random sampling lets us surface problems without that knowledge, suggesting that the dataset-creation pipeline as a whole produces them, not just isolated cases. Third, prior work focused mostly on uncurated, web-scraped datasets, whereas we also audit a highly curated dataset (Framed Multi30K) and a filtered one (relaion2B-en-research-safe), showing that bias and ethical concerns persist after careful curation or automated safety filtering. Fourth, automated face and sentiment classifiers bring their own biases into the analysis~\cite{dominguez2026unmasking, birhane2023into}. We instead use standardized AIETs to guide a manual audit, discussing findings with a multidisciplinary group to capture problems that machines may miss.

Our core argument is that standard dataset documentation AIETs can identify bias and ethical problems using a very small sample. This works because most datasets never went through basic human curation. Since the original scraping and collection processes usually lack human participation, a qualitative review of a small sample is enough to reveal these problems, without high-end computing or complex algorithmic filters. We make no claim to exhaustiveness: because knowledge is always situated~\cite{haraway1988situatedknowledges}, no audit can identify every ethical problem in a dataset, and that is not the aim of this paper. Detecting such problems does not require comprehensive review. As the adage holds, \textit{you don’t need to drink the whole ocean to know it’s salty}.

\section{Method}
\label{sec:method}

In this section, we detail our proposed methodology for auditing massive datasets through small random samples. The pipeline has five steps: (1) selecting a dataset and reviewing an external documentation (Section~\ref{subsec:dataselec}), (2) drawing a random sampling strategy (Section~\ref{subsec:randomsamp}), (3) choosing an appropriate AIETs (Section~\ref{subsec:defaiet}), (4)~applying the AIET through documentation review combined with manual and automated qualitative sample analysis (Section~\ref{subsec:application}), and (5)~incorporating multidisciplinary analysis to ensure a comprehensive ethical audit guided by diverse perspectives and backgrounds~(Section~\ref{subsec:analysis}). 

\subsection{Dataset Selection and Documentation Review}
\label{subsec:dataselec}
The auditing process begins by selecting a target dataset to analyze for bias, technical flaws, and ethical concerns. Selection requires only one condition: access to at least a subset of the dataset, not the full dataset. This process applies to datasets that have already been published or datasets currently in development or curation.

Once the dataset is selected, auditors conduct an initial review of external content and documentation, such as papers, previous documentations, and repositories. Reviewing these materials familiarizes the auditors with the dataset's creators, motivations, and curation processes. It also provides the context needed to answer questions that data inspection alone cannot be address. 

\subsection{Random Sampling}
\label{subsec:randomsamp}

After reviewing the documentation, auditors select a random sample of the dataset. The appropriate sample size  depends on data characteristics and audit team capacity. High annotation density requires smaller sample sizes to maintain rigorous evaluation within time constraints, while unannotated data allows larger sample sizes. Additionally, audit capacity increases with team size and domain familiarity, allowing larger sample sizes relative to smaller or less experienced groups. In this paper, we audited 1\% sample of the curated \textit{Framed Multi30K} dataset~\cite{viridiano2024framed}, and 0.00006\% of the uncurated \textit{relaion2B-en-research-safe} dataset~\cite{relaion}.

\begin{figure*}[t]
\centering
\includegraphics[width=\textwidth]{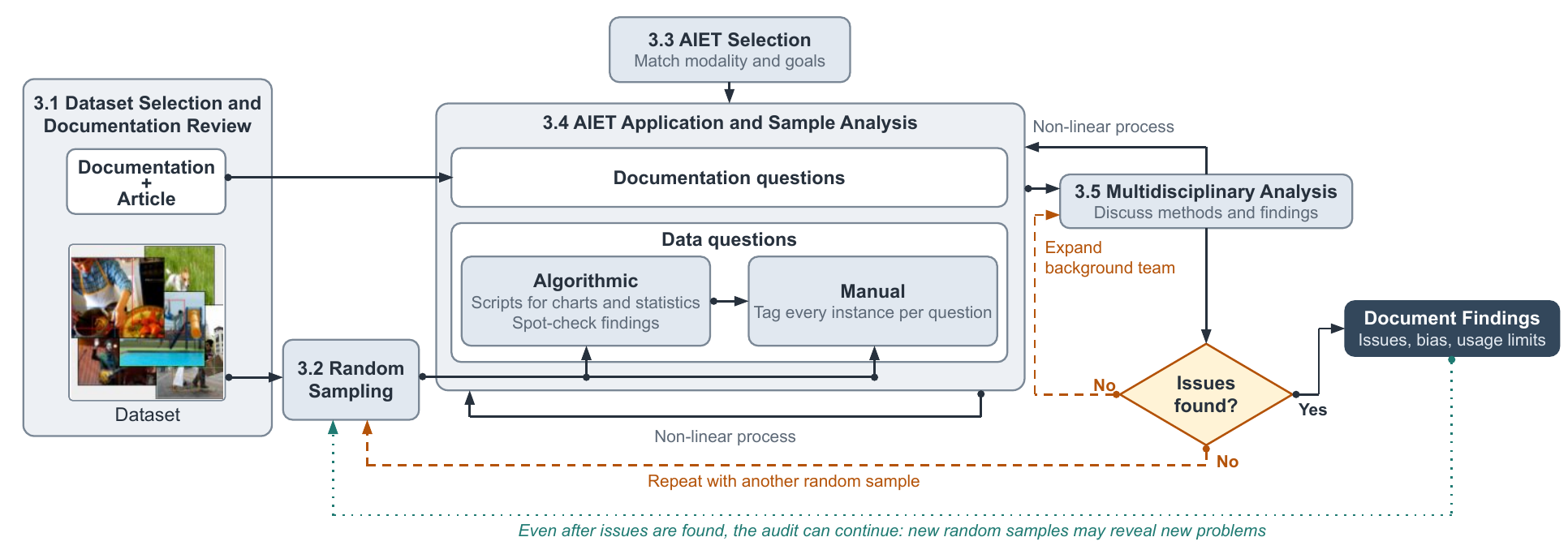}
\caption{Proposed pipeline.}
\label{fig:imagempipeline}
\end{figure*}

\subsection{AIET Selection}
\label{subsec:defaiet}

Auditors must select an AIET aligned with their specific data modalities, structural formats, and analytical objectives. Given the heterogeneity of AIETs across dataset lifecycle and focus area, auditors should consult domain literature and surveys, such as~\citet{reynolds2026reflection}, to identify tools better suited for their target dataset and~goals. In this paper, we selected four distinct AIETs:  \textit{Datasheets for Datasets}~\cite{gebru2021datasheets}, \textit{Data Statements}~\cite{bender2018data,10.1145/3594737}, \textit{Dataset Nutrition Label}~\cite{chmielinski2022dataset, holland2018}, and \textit{Data Cards}~\cite{pushkarna2022data}.

\subsection{AIET Application and Sample Analysis}
\label{subsec:application}

Auditors apply the chosen AIET by addressing the tool's questions and evaluating the dataset sample alongside supporting documentation. 
The application process is non-linear and iterative: answering later questions often uncovers new perspectives that encourages auditors to revise and refine earlier responses. To structure this evaluation, the tool's questions are divided into two categories:

\begin{enumerate}
    \item \textbf{Documentation questions}. 
Questions about dataset characteristics such as motivation, collection process, maintenance, and distribution, are answered directly using the previously collected external sources without requiring data inspection. In applied AIETs, these correspond to specific questions:
\begin{itemize}
    \item \textit{Datasheets for Datasets}: The \textit{Motivation}, \textit{Collection process}, \textit{Preprocessing/cleaning/labeling}, \textit{Distribution}, and \textit{Maintenance} sections; questions 7 and 12 in \textit{Composition}; and all questions except 40 in \textit{Uses}.
    \item \textit{Data Statements}: The \textit{Header}, \textit{Curation rationale}, \textit{Documentation for source datasets}, \textit{Language varieties}, \textit{Annotator demographic}, \textit{Preprocessing and data formatting}, \textit{Metadata}, \textit{Disclosures and ethical review}, \textit{Distribution}, and \textit{Maintenance} sections. 
    \item \textit{Data Nutrition Label}: The \textit{Metadata} section (excluding questions 1 and 2); \textit{Just facts} (excluding 15 and 25); \textit{Collected (Why)} (excluding 32, 34, 35, 36, and 37); question 40~in \textit{Collected (What)}; \textit{Collected (How)} (excluding 45); \textit{Processed (How)} (excluding 48); and question 50 in \textit{Known issues}.
    \item \textit{Data Cards}: The \textit{Authorship}, \textit{Motivations \& intentions}, \textit{Access, retention, \& wipeout}, \textit{Provenance}, \textit{Extended use}, \textit{Transformations}, \textit{Annotations \& labeling}, \textit{Validation types}, \textit{Sampling methods}, and \textit{Known applications \& benchmarks} sections, as well as Dataset Version and Maintenance within \textit{Dataset overview}.\vspace{0.1cm}
\end{itemize}

\item \textbf{Data questions}. 
Questions related to dataset composition, instances, and representation require direct empirical analysis of the data. In applied AIETs, these correspond to specific sections:
\begin{itemize}
    \item \textit{Datasheets for Datasets}: The \textit{Composition} section (excluding questions 7 and 12) and question 40 in \textit{Uses}. 
    \item \textit{Data Statements}: The \textit{Executive summary}, \textit{Language varieties}, \textit{Language user demographic}, \textit{Linguistic situation and text characteristics}, \textit{Capture quality}, \textit{Limitations} and \textit{Other} sections. 
    \item \textit{Data Nutrition Label}: Questions 15 and 25 in \textit{Just facts}; questions 32, 34, 35, 36, and 37 in \textit{Collected (Why)}; the \textit{Collected (What)}; questions 44 and 45 in \textit{Collected (How)}; question 48 in \textit{Processed (How)}; and question~51 in \textit{Known issues}. 
    \item \textit{Data Cards}: The \textit{Summary}, \textit{Dataset overview} (excluding \textit{Dataset version and maintenance}), \textit{Example of data points}, \textit{Human and other sensitive attributes}, \textit{Sampling methods}, and \textit{Reflections on data} sections.\vspace{0.1cm} 
\end{itemize}

Depending on the question scope and the existing annotation density, dataset evaluation proceeds manually, algorithmically, or through a hybrid approach. 

Questions about macro-level patterns, label distribution, or term frequencies are initially processed with automated scripts when the dataset structure allows. Automated extraction is often complemented by manual analysis of specific instances, images, and their captions. 
For example, when analyzing subpopulations in annotated datasets (e.g.,  Framed Multi30K), auditors algorithmically analyze subpopulation distributions across captions and languages prior to manual inspection. Conversely, in unannotated datasets (e.g., relaion2B-en-research-safe), auditors must manually annotate subpopulation occurrence before any analysis.

Questions requiring visual, semantic, or contextual analysis necessitate qualitative manual analysis, where auditors inspect and annotate each instance according to the question. Visualization tools such as FiftyOne~\cite{moore2020fiftyone} facilitate this stage by enabling interactive data inspection, allowing auditors examine images with associated bounding boxes, annotations, metadata, and custom annotations.
\end{enumerate}
Through this data evaluation, every sample instance undergoes human qualitative review, helping identify systemic bias, issues, and ethical concerns that automated methods fail to detect.

\subsection{Multidisciplinary Analysis}
\label{subsec:analysis}
Comprehensive ethical auditing requires evaluating methods and findings within a multidisciplinary team, ensuring a diverse repertoire of worldviews, opinions, and backgrounds. 

When specific questions, particularly regarding data collection, dissemination, or maintenance, cannot be answered directly using available materials, direct communication with  dataset creators may be required. 

Critically, failing to detect biases does not guarantee that the dataset is flawless. If auditors find no issues, they should either draw an additional random sample or involve auditors with different contextual, educational, and professional backgrounds to uncover previously unnoticed issues.

Fig.~\ref{fig:imagempipeline} illustrates the workflow for applying an AIET tool to a random sample of a target dataset.

\section{Results and Discussion}
\label{sec:results}

In this section, we present the empirical evaluation of two dataset case studies (\textit{Framed Multi30K}~\cite{viridiano2024framed} and \textit{relaion2B-en-research-safe} \cite{relaion}) using the proposed AIET auditing framework. In Section~\ref{subsec:datasets}, we detail the selected datasets. In Section~\ref{subsec:results}, we analyze the audit findings regarding biases, privacy, and ethical concerns, followed by a discussion of how our framework can be generalized across diverse datasets.

\subsection{Datasets}
\label{subsec:datasets}

\subsubsection{\textbf{Framed Multi30K}}
\textit{Framed Multi30K}~\cite{viridiano2024framed} is an open, multimodal, and multilingual dataset that extends the benchmark \textit{Flickr\-30K}~\cite{young-etal-2014-image}, \textit{Multi30K}~\cite{elliott2016multi30k}, and \textit{Flickr30K Entities}~\cite{plummer2015flickr30k} datasets by incorporating FrameNet-based fine-grained semantic representations. It integrates, via automated text annotation and human image annotation, schematic knowledge structures called frames and frame elements to describe situations, objects, events, and entities, detailing the participants and conceptual roles necessary for their representation~\cite{10.3389/fpsyg.2022.838441}.

The dataset significantly expands linguistic diversity by introducing Brazilian Portuguese into the Multi30K family. It incorporates 158,915 original Brazilian Portuguese descriptions produced independently by native speakers, alongside 31,104 Brazilian Portuguese translations aligned with the original English descriptions and existing German, French, and Czech translations. Furthermore, the corpus adds 4,577,122 automatic frame and frame element labels to both the English and Brazilian Portuguese descriptions using \textit{Lome}~\cite{xia2021lome}, a multilingual semantic parser trained on  Berkeley FrameNet 1.7~\cite{framenet2009} and FrameNet Brasil~\cite{10.3389/fpsyg.2022.838441}.

To enrich visual grounding, the dataset extends the \textit{Flickr30K Entities} by manually  annotating 29,920 sets of bounding boxes using the multimodal annotation tool \textit{Charon}~\cite{belcavello2022charon}. This  yielded 169,560 correlations between frames, frame elements, and pre-existing phrase-to-region configurations. The annotations were produced under two experimental conditions: with the image descriptions available as context (visual with context -- VWC) and without them (visual no context -- VNC).

For our audit of \textit{Framed Multi30K}, we created a random sample of 1\%
, composed of 316 images, 1,580 original English captions, 1,580 original Brazilian Portuguese captions, and 306 translated Brazilian Portuguese captions.

\begin{figure*}[ht]
\begin{subfigure}{0.49\textwidth}
    \hspace{0.25cm}
    \includegraphics[width=0.415\linewidth,clip,trim={0 3cm 0 0}]{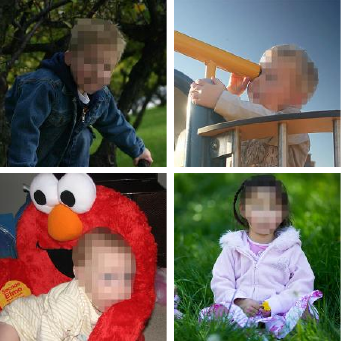}
    \includegraphics[width=0.415\linewidth,clip,trim={0 0 0 3cm}]{imagens/criancas.pdf}
    \caption{Children}
    \label{fig:imagemCrianca}
\end{subfigure}
\begin{subfigure}{0.5\textwidth}
    \includegraphics[width=0.5\linewidth,clip,trim={0 10cm 0 0}]{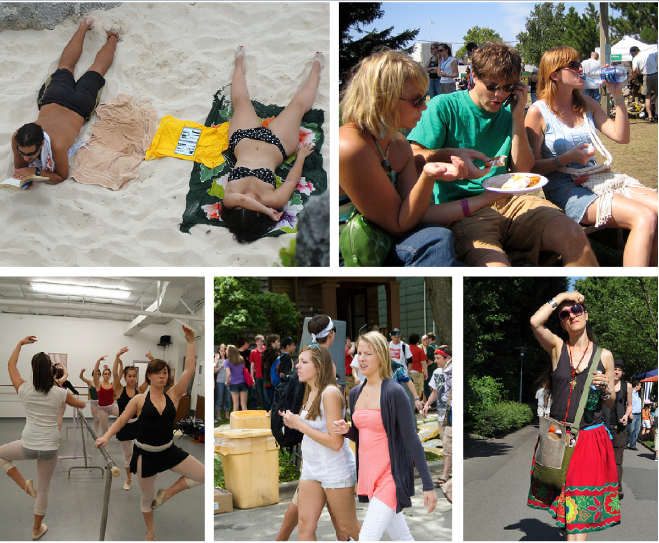}
    \includegraphics[width=0.5\linewidth,clip,trim={0 0 0 10cm}]{imagens/menina_girl.png}
    \caption{Women}
    \label{fig:legendaMulher}
\end{subfigure}
\caption{Example: (a) identifiable children and (b) women described as ``menina'' or ``girl''.\vspace{0.15cm}}
\label{fig:figuraExemplos}
\end{figure*}

\subsubsection{\textbf{relaion2B-en-research-safe}}
\textit{\hspace{0.2cm}relaion2B-en-research-safe}\\\cite{relaion} is an open, web-scale text-to-image dataset designed for langua\-ge-vision learning, revised to remove known links to suspected Child Sexual Abuse Material (CSAM) and Not Safe For Work (NSFW) content. 
It represents a safety-revision subset of the original \textit{LAION-5B} dataset~\cite{schuhmann2022laion}, which comprised roughly 5.85 billion image-text pairs. After a December 2023 report identified CSAM links~\cite{thiel2023identifying}, LAION partnered with the Internet Watch Foundation, the Canadian Center for Child Protection, and Stanford to remove 2,236 suspected CSAM links. Following Human Rights Watch guidance, researchers also removed an additional 399 links exposing children's private information. 

The updated dataset comprises two subsets based on a CLIP-based NSFW detector~\cite{clip_based_nsfw_detector} and text keyword filters. The \textit{research} version applies a threshold filter of \textit{p\_unsafe} $> 0.95$,  removing 1.121\% (approximately 65~million) of original samples. The more restrictive \textit{research-safe} version enforces a threshold of \textit{p\_unsafe} $ > 0.45$, removing 3.044\% (approximately 176 million) of samples from the original dataset.

For our audit of the \textit{relaion2B-en-research-safe} dataset, we extracted a 0.00006\% sample 
 comprising 1,200 image-text pairs alongside their corresponding metadata attributes:  \textit{p\_unsafe} (NSFW probability), 
\textit{p\_wa\-termark} (watermark probability), and \textit{similarity} (image-caption similarity). The sample size was deliberately constrained to reflect the practical capacity of individual researchers to manually inspect and document data, rather than relying on large auditing teams or extensive computational resources. This choice reflects the central premise of our study: that meaningful ethical audits of massive datasets can be conducted by a small number of human auditors using accessible AIETs. We created two non-overlapping random samples of this percentage.

\subsection{Results}
\label{subsec:results}

\subsubsection{\textbf{Framed Multi30K}}

The application of the tools revealed issues regarding the dataset's composition, both inherited from the \textit{Flickr30k} dataset family~\cite{young-etal-2014-image} (\textit{Multi30k}~\cite{elliott2016multi30k} and \textit{Flickr30k Entities}~\cite{plummer2015flickr30k}), such as the lack of \textit{bounding boxes} in 13 images from \textit{Flickr30k Entities} and the absence of translated descriptions for 10~images in \textit{Multi30k} and issues originating directly from \textit{Framed Multi30k}.  Beyond the detailed description of dataset biases in \citet{Viridiano2024FM30K}, direct contact with the authors of the \textit{Framed Multi30k} dataset helped us identify these issues and their origins, as well as other problems concerning its integrity. 

As a consequence of expanding the \textit{Flickr30K} family by \textit{Framed Multi30K}, we identified ethical concerns regarding surveillance, minors' privacy, and the public exposure of their images in computer vision models, with approximately 14\% of the images containing identifiable children, as illustrated in Fig.~\ref{fig:imagemCrianca}. Similarly, we observed the presence of gender-related linguistic bias in adult women when frequently described by the term ``\textit{girl}'', derived from \textit{Flickr30K} captions, or ``menina'', coming directly from \textit{Framed Multi30k} through captions translated from English or originally produced by Brazilian annotators. Fig.~\ref{fig:legendaMulher} presents some examples of women described in such way. These choices perpetuate sensitive and potentially offensive stereotypes that models trained on such data can amplify.

Furthermore, the dataset reflects representation biases due to its image source (everyday scenes from Flickr), with underrepresentation of non-urban, non-Western, and low-resource cultural contexts, and overrepresentation of perspectives from highly educated Brazilian Portuguese speakers. Certain activities, social roles, and visual contexts are more frequent than others, which may limit how well the dataset represents broader global or demographic populations. 

Regarding the identification and representation of subpopulations in the data, the analysis of \textit{frames} automatically annotated by the \textit{Lome} model --- especially concerning the association of attributes, leisure activities, vocation, religion, age, ethnicity, jurisdiction, morality, origin, transitory activity, and transport mode to individuals mentioned in the captions 
--- revealed systematic differences between Portuguese and English. In Portuguese, terms like ``branco'' (white) and ``negro'' (black) are more frequently annotated by the model as ethnicity, whereas in English, markings for Asian ethnicity and compound designations such as \textit{African American} and \textit{Native American} predominate. This is due to the fact that, during training, \textit{Lome} is exposed to human-made annotations in both languages: Portuguese and English. While Portuguese annotations come from the \textit{FrameNet Brasil} dataset~\cite{10.3389/fpsyg.2022.838441}, the English ones derive from \textit{Berkeley FrameNet}~\cite{framenet2009}. 

\begin{figure*}[ht]
    \subfloat[Image]{\includegraphics[width=0.25\linewidth]{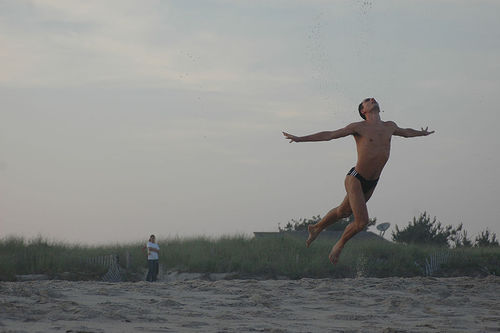}}
    \hspace{0.35cm}
    \subfloat[Corresponding caption]{\includegraphics[width=0.55\linewidth,clip,trim={0 0 2cm 0}]{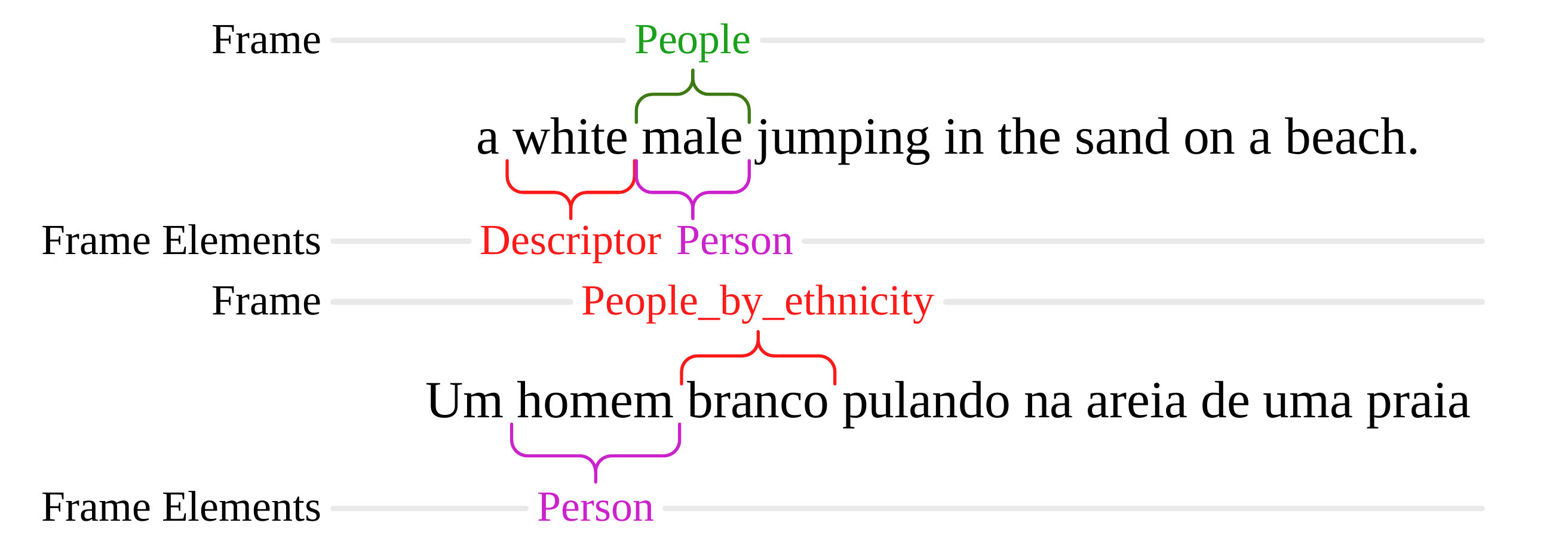}}
\caption{Example: (a) original image and (b) associated \mbox{description.}\vspace{0.155cm}}
\label{fig:figura-branco}
\end{figure*}

A query to these datasets shows that, while the Brazilian dataset has the form ``branco'' associated with and annotated for \textit{frames} related to color and ethnicity, no sentences containing the form ``\textit{white}'' are annotated with ethnicity-related \textit{frames} in the \textit{Berkeley FrameNet} dataset. This suggests that \textit{Lome} may not have encountered this specific annotation pattern in its English training data.  Consistent with this observation, Fig.~\ref{fig:figura-branco} shows that the word ``\textit{white}'' was automatically annotated as a ``descriptor'' of a person. However, in the same sentence translated to Portuguese, ``branco'' was automatically annotated with a label related to ethnicity. 

\begin{figure*}[ht]
    \subfloat[Image]{\includegraphics[width=0.25\linewidth]{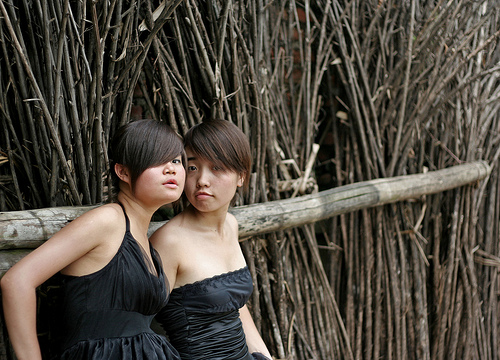}\label{fig:imagemAsiatico}}
    \hspace{0.5cm}
    \subfloat[Corresponding caption]{\includegraphics[width=0.6\linewidth]{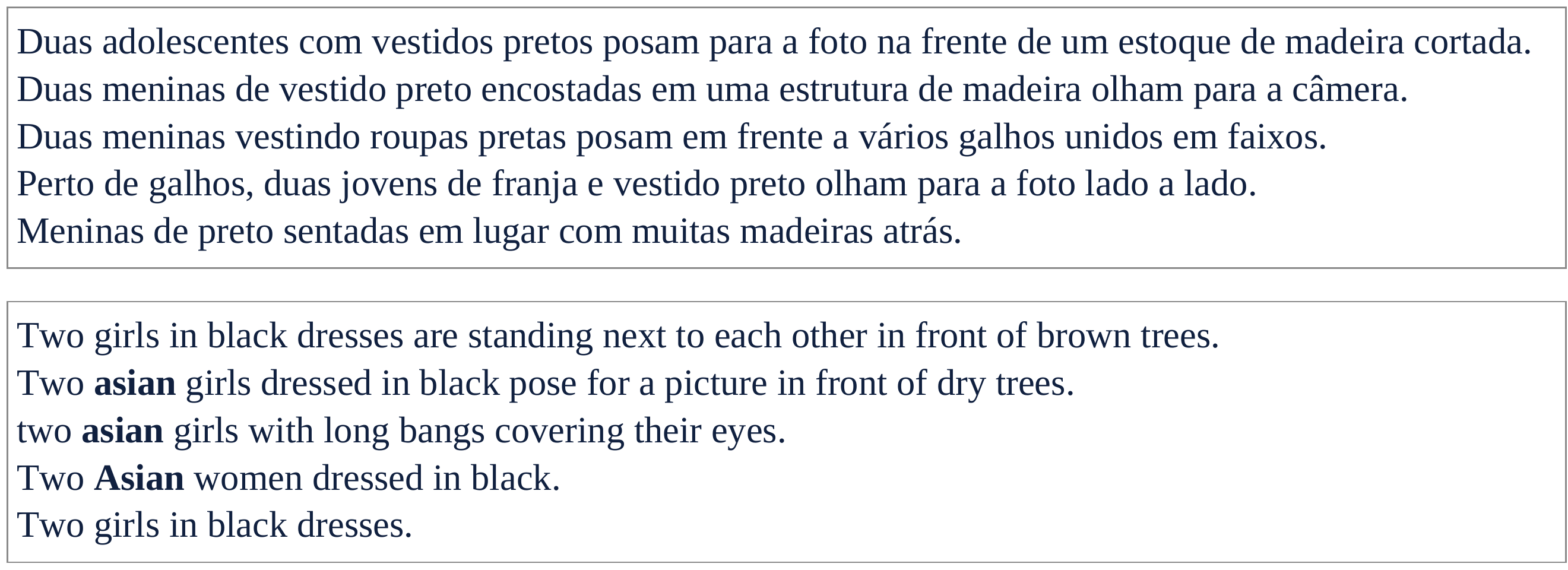}\label{fig:legendaAsiatico}}
\caption{Example of an image and its respective visual caption: (a) original image and (b) associated description.\vspace{0.15cm}}
\label{fig:figuraAsiatico}
\end{figure*}

In contrast, human annotators mark Asian ethnicity much more in English than in Portuguese, when creating the image descriptions. In Fig. \ref{fig:figuraAsiatico}, of the 5 English captions, 3 mark the people in the image as Asian, while none of the Portuguese captions mention their ethnicity.

We also identified qualitative issues in the automatic annotations. Regarding the ``black'' ethnicity, we observed that the term is more frequently annotated in Portuguese than in English, as a reflection of usage patterns in both languages. English captions more often use the ``African'' or ``African American'' nationalities, while Portuguese captions tend to use terms translated as ``black''. Moreover, one instance showed the ethnicity ``\textit{black}'' automatically annotated as a ``color'' in English, rather than a label related to ethnicity or origin, not representing the broader social and contextual meaning associated with the term.

Similarly, the term ``American'' appeared exclusively in English captions and was frequently automatically annotated by LOME with \textit{frames} related to origin. The presence of this term in the captions suggests that crowd workers involved in the creation of the original Flickr 30k dataset inferred people's nationality solely from visual information, which is insufficient for a correct inference~\cite{van2016stereotyping}. However, the same does not hold for the Brazilian Portuguese captions created by expert annotators. Furthermore, the term did not appear in isolation, but in compounds such as ``\textit{Native American}'' and ``\textit{African American}''. These cases highlight limitations and inconsistencies in the annotation process and reflect limitations inherited from the original description creation process by \textit{crowd workers} of \textit{Flickr30K}.

Thus, these patterns evidence linguistic and cultural biases in how the original dataset identifies subpopulations, with a tendency to mark or annotate only ethnicities and origins considered non-standard from the point of view of the crowd workers involved in the creation of Flickr30k, reinforcing the concept of ``white'' as the non-racialized default. However, by looking at each image, the vast majority of people were white~\cite{van2016stereotyping}.

Furthermore, the dataset has linguistic representation limitations in the captions originally created for Brazilian Portuguese regarding regional dialects and register. The Brazilian Portuguese descriptions are written in a specific linguistic variety spoken by highly educated, urban speakers of the state of Minas Gerais in monitored communicative settings, introducing a socio-educational \mbox{language-variety bias.}

As for the human semantic annotation of images, the analysis shows a somewhat different scenario. Concerning the representation of people, among the 62 images annotated for the people\_by\_age frame in the two annotation conditions, only 4 do not portrait children or elderly people. Such a pattern reinforces the tendency of annotators to mark characteristics that are different from their own, as all annotators are young adults. The VWC annotation condition corresponds to 57 instances of the frame, while the VNC condition responds to 43, indicating the reinforcement of eventual biases by the original English caption.

For the people\_by\_gender frame, human annotations are more diverse but focus mostly on adult individuals. However, some images have individuals annotated for this frame which are shown in a way that is not possible to assign any sort of gender annotation. Once again, annotation condition seems to play an important role here: among all the instances of this frames, 43 were annotated under the VWC condition, while 29 refer to the VNC condition. Moreover, in those last 29 annotations, very few are grounded on individuals that were photographed at a distance or from the back.

The findings across all four documentation tools were very similar to each other. Thus, we decided to use only \textit{Datasheets for Datasets} for \textit{relaion2B-en-research-safe}, but analyze more than one sample, since our sample size is extremely small compared to the dataset.

\subsubsection{\textbf{relaion2B-en-research-safe}}

\begin{figure*}[tbp]
    \centering
    
    \subfloat[]{\includegraphics[height=2cm,clip,trim={6cm 0 6cm 0}]{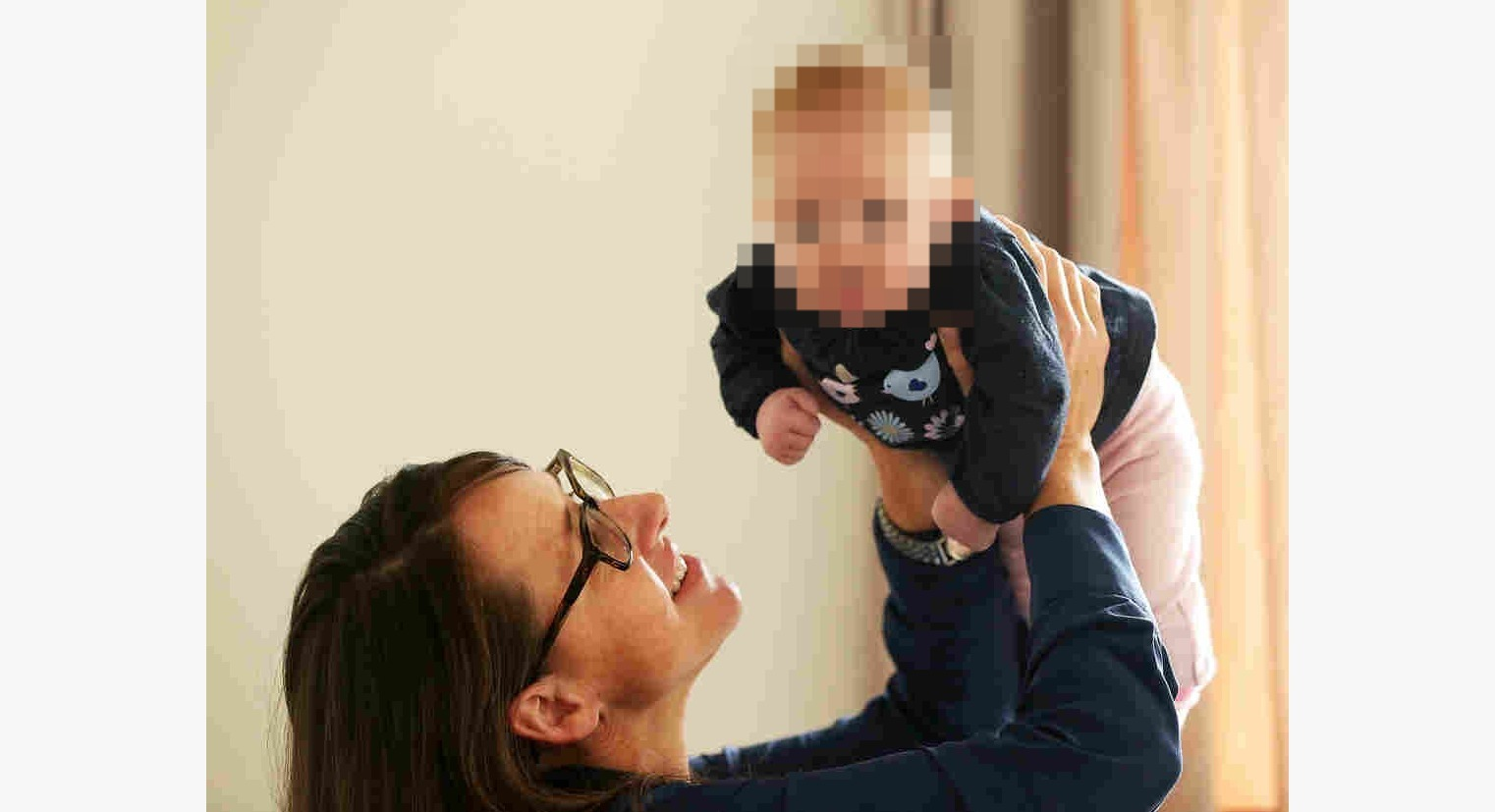}\label{fig:mother}}
    \subfloat[]{\includegraphics[height=2cm,clip,trim={0cm 0 0cm 0}]{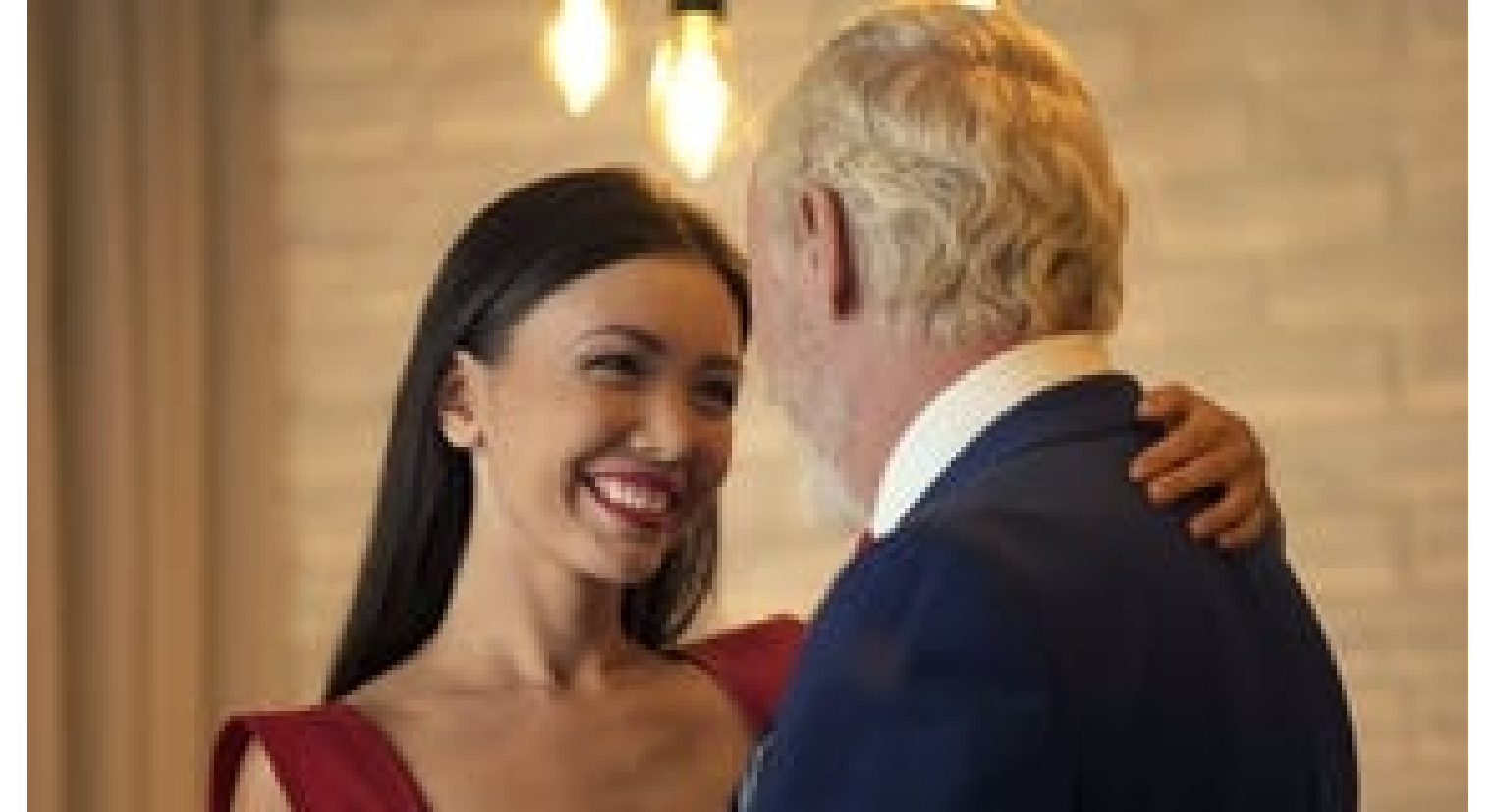}\label{fig:flirting}}
    \subfloat[]{\includegraphics[height=2cm,clip,trim={0 0 0cm 0}]{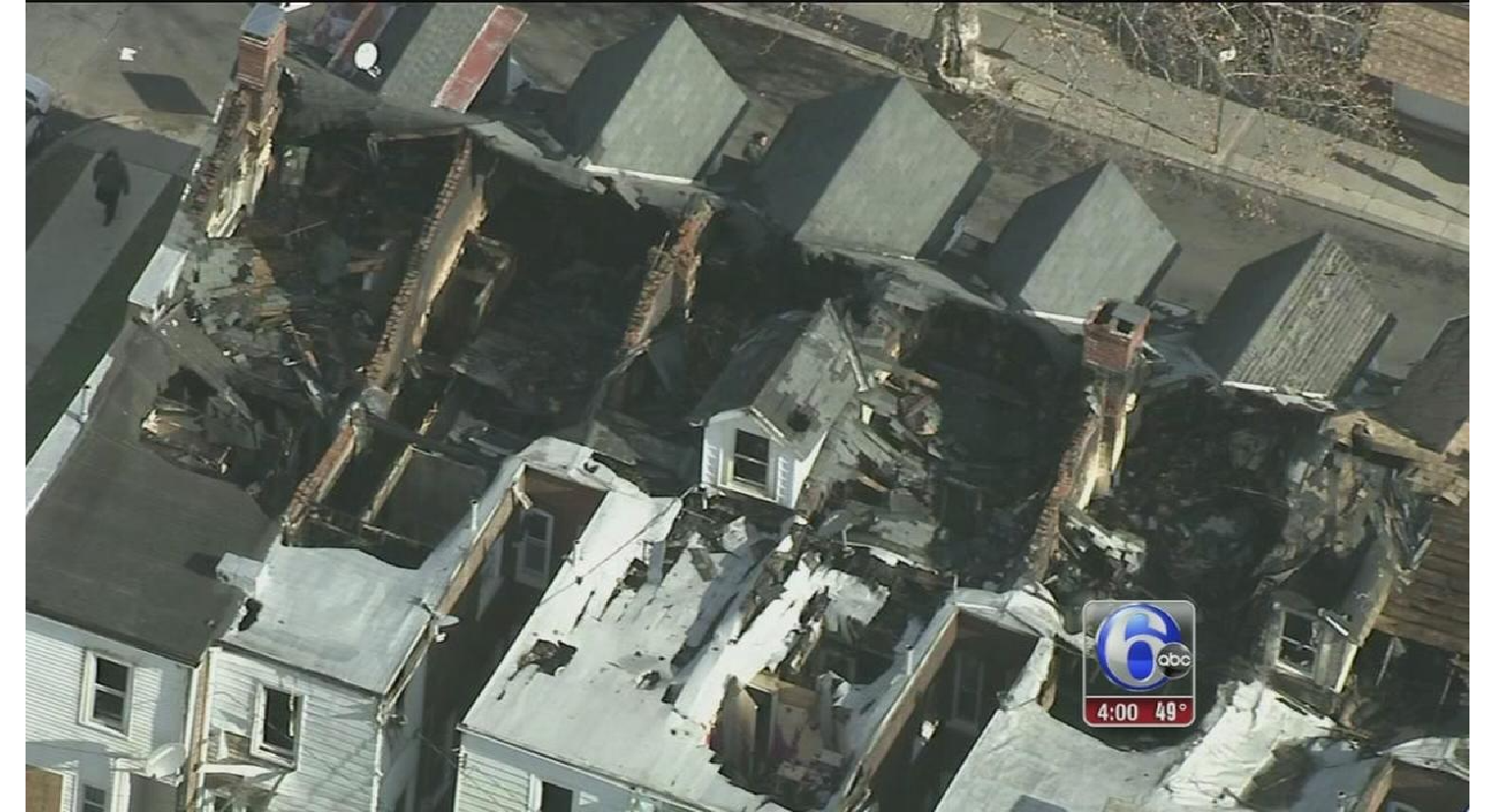}\label{fig:house}}
    \subfloat[]{\includegraphics[height=2cm,clip,trim={6cm 0 6cm 0}]{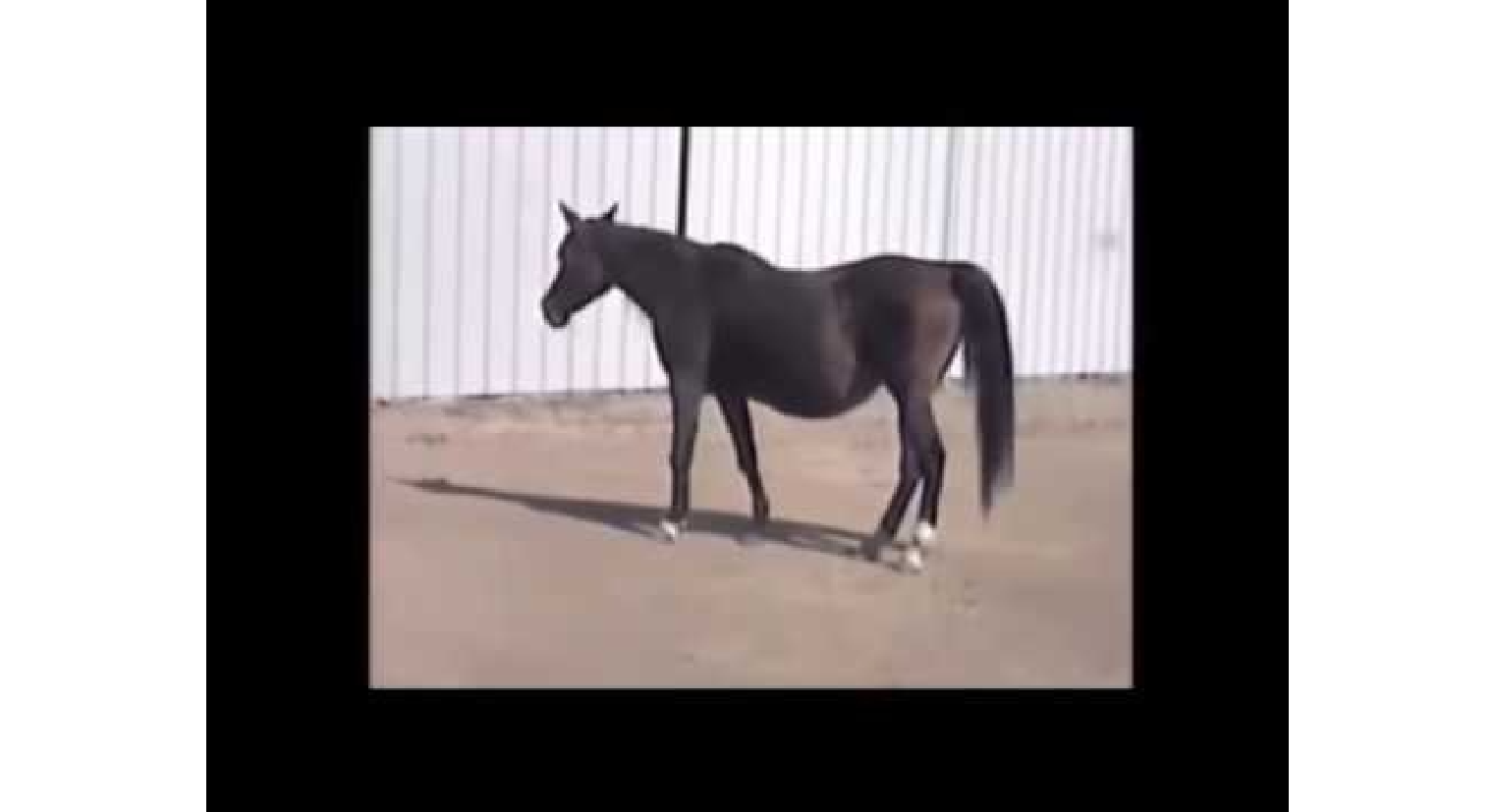}\label{fig:horse}}
    \subfloat[]{\includegraphics[height=2cm,clip,trim={16cm 0 16cm 0}]{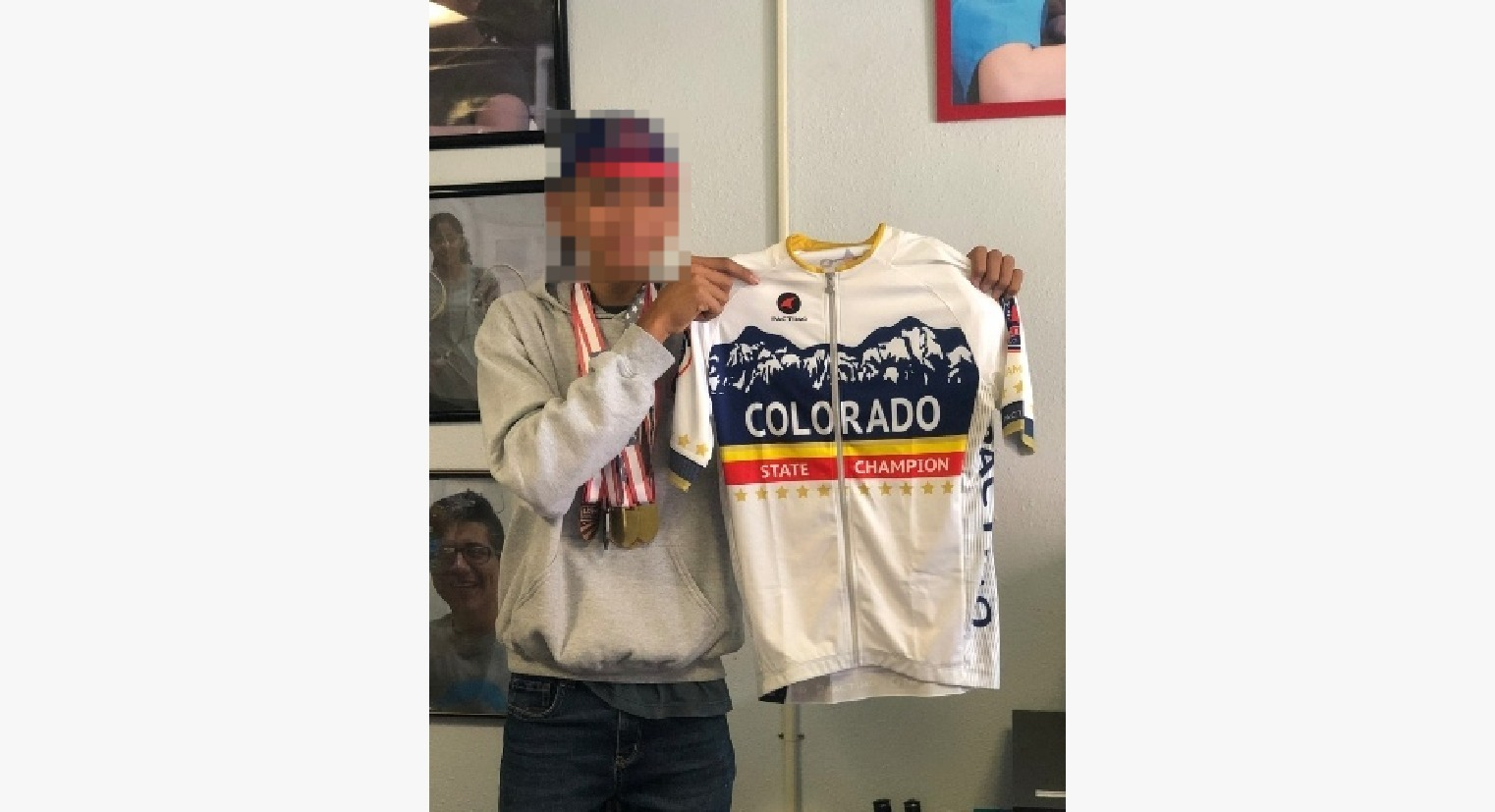}\label{fig:children1}}\vspace{0.1cm}\\
    \subfloat[]{\includegraphics[height=2cm,clip,trim={6cm 0 6cm 0}]{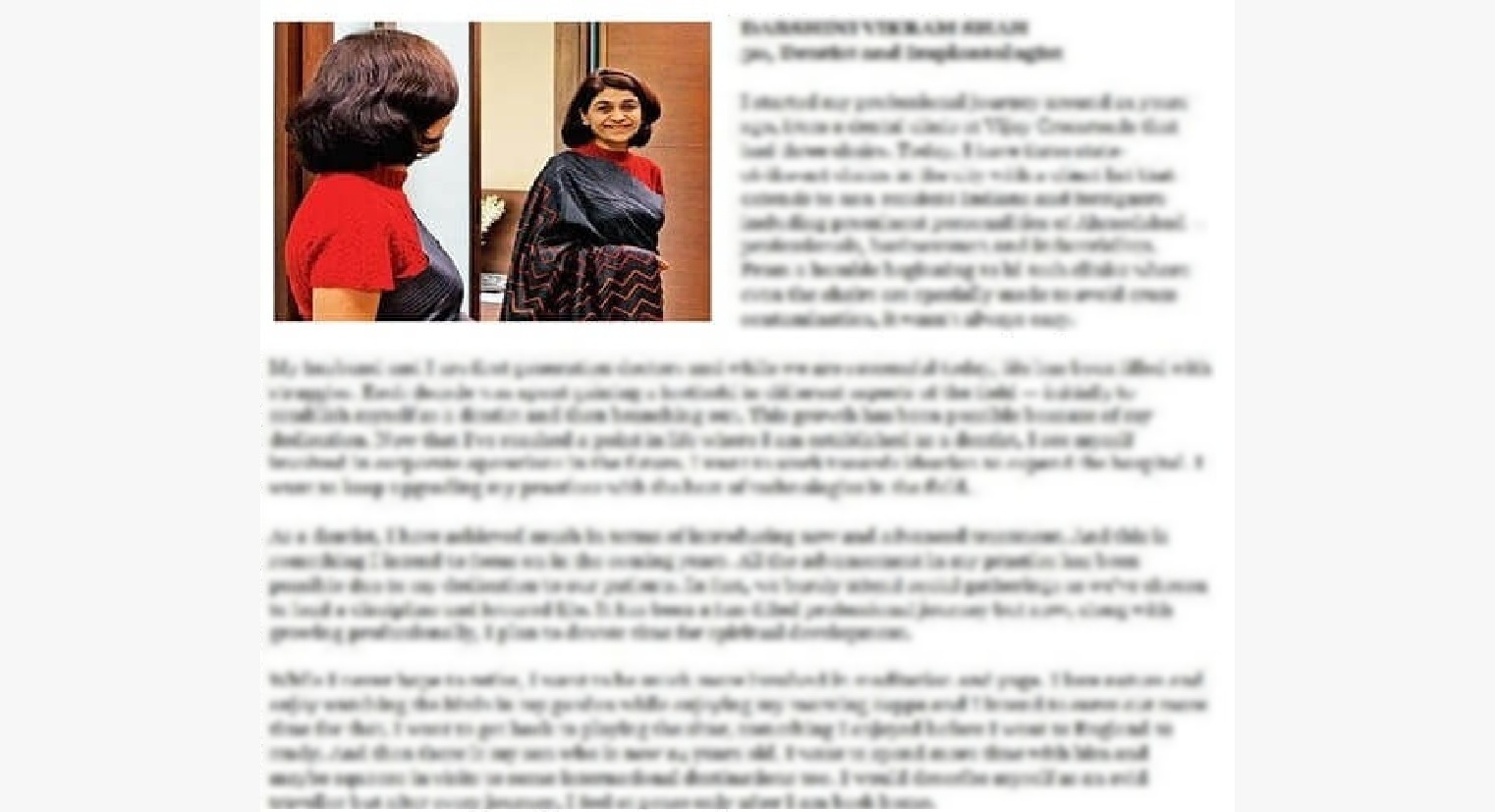}\label{fig:text}}
    \subfloat[]{\includegraphics[height=2cm,clip,trim={0cm 0 0cm 0}]{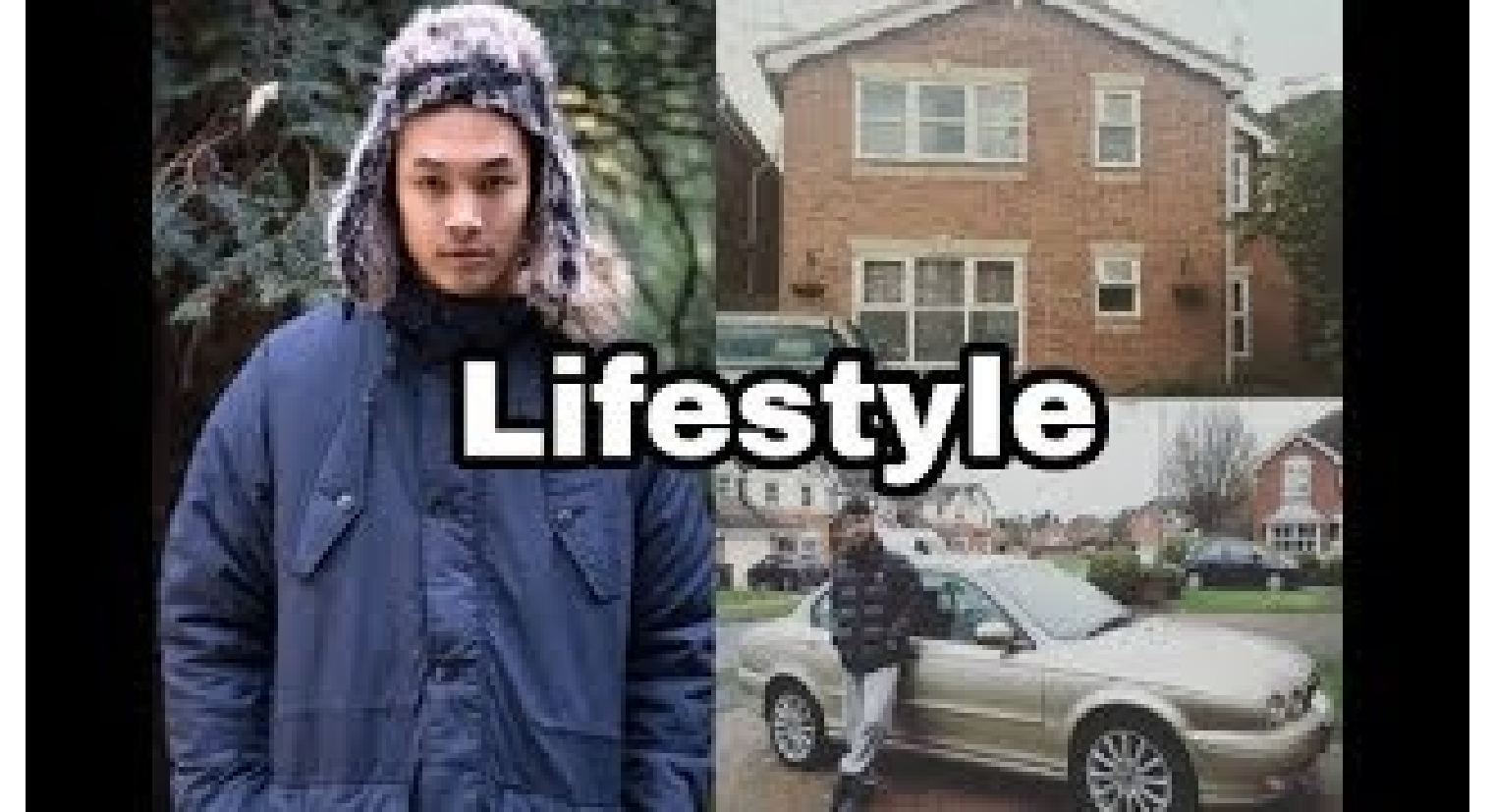}\label{fig:nationality}}
    \subfloat[]{\includegraphics[height=2cm,clip,trim={6cm 0 6cm 0}]{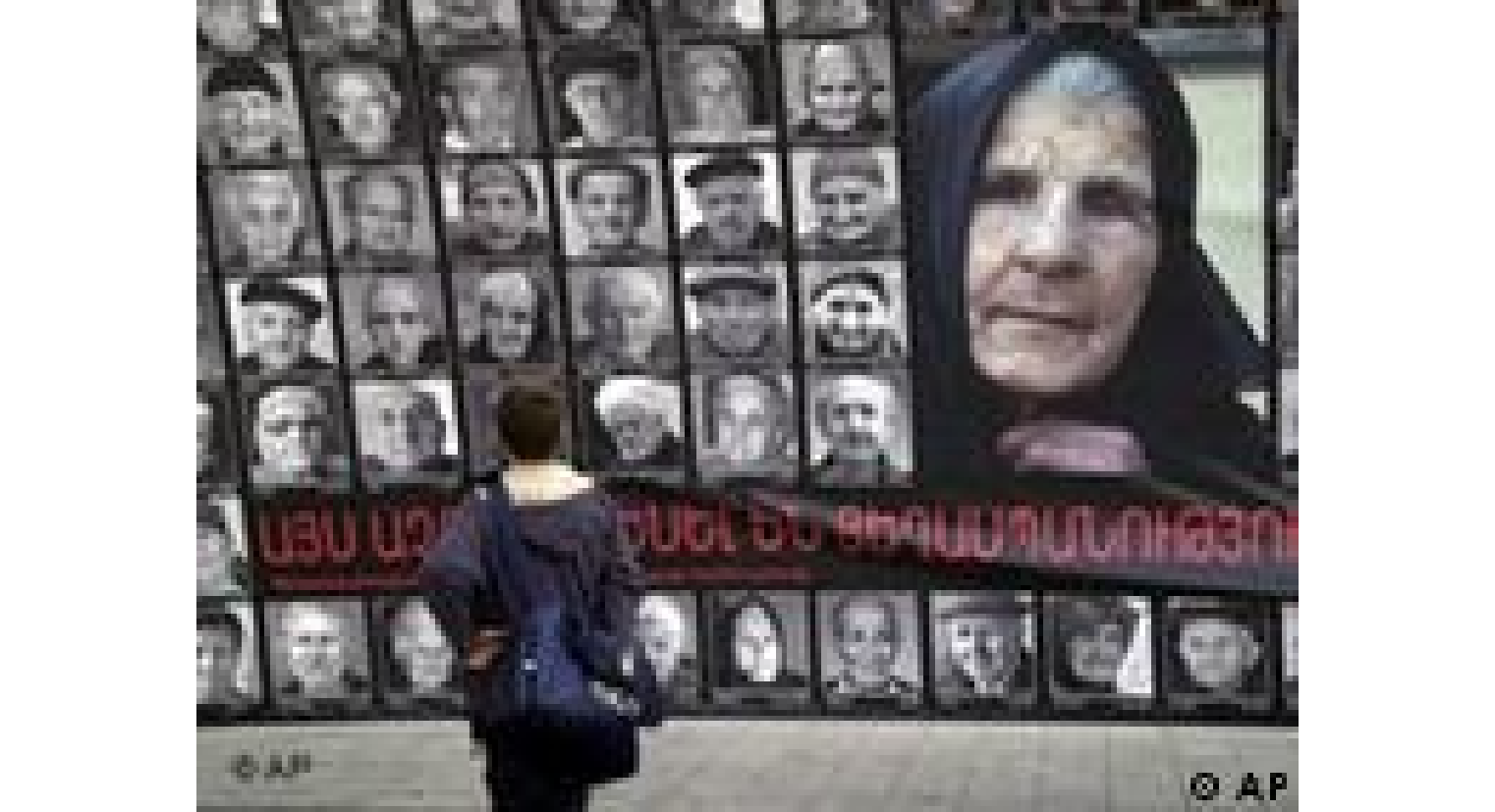}\label{fig:portraits}}
    \subfloat[]{\includegraphics[height=2cm,clip,trim={16cm 0 16cm 0}]{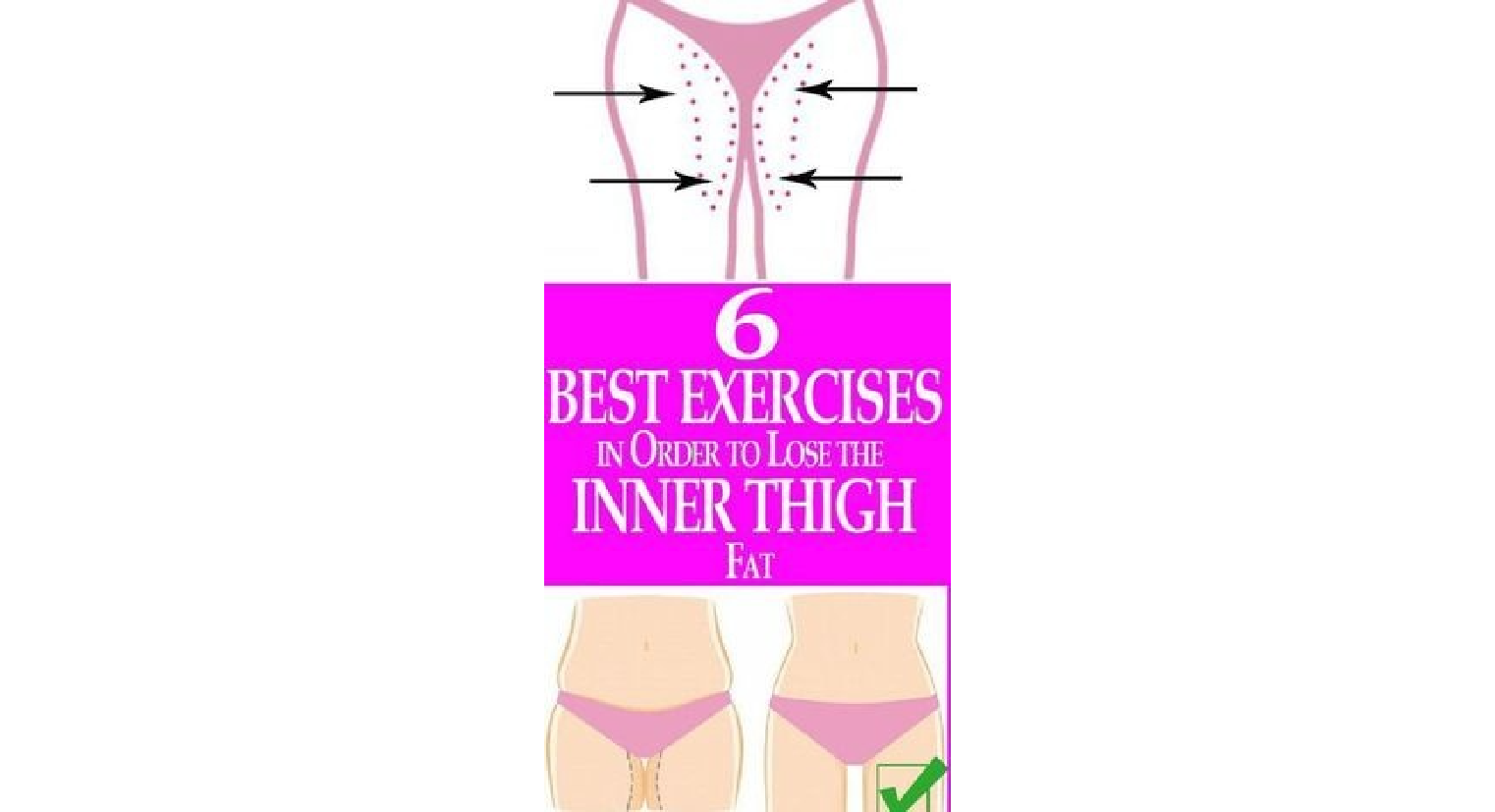}\label{fig:weight}}
    \subfloat[]{\includegraphics[height=2cm,clip,trim={12cm 0 12cm 0}]{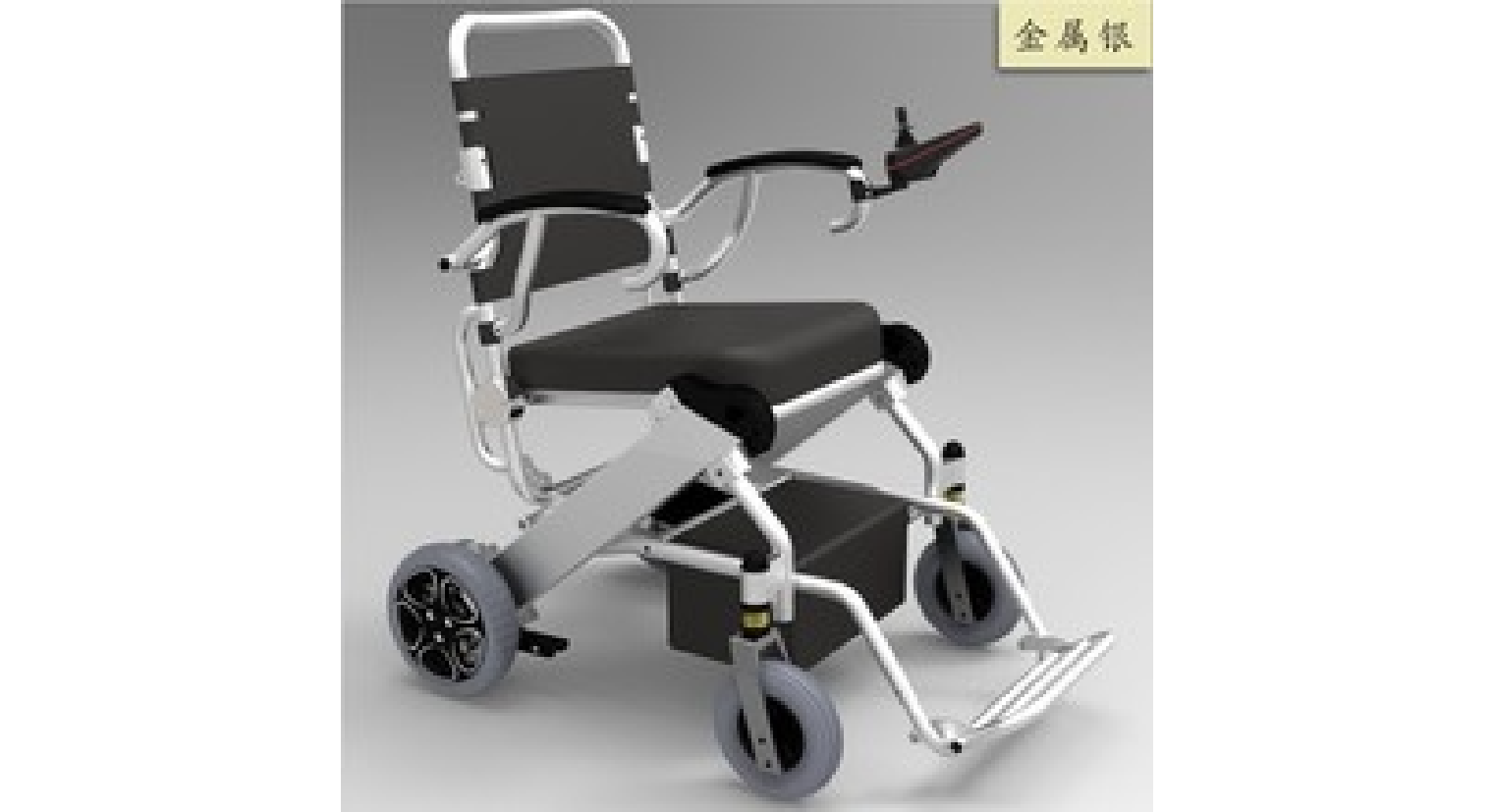}\label{fig:chair}}
    \caption{\textbf{Sample instances from \textit{relaion2B-en-research-safe}.} 
    \textbf{(a)}~Secondary school teacher [X] at home on maternity leave with her 3-month-old daughter, [X]. A resident of New Zealand, [X] has access to paid leave, something many American mothers do not. 
    \textbf{(b)}~Young Woman Flirting And Dancing With Old Millionaire On Date, Escort Service Stock Photos. 
    \textbf{(c)}~VIDEO: 6 homes destroyed in massive Allentown blaze. 
    \textbf{(d)}~Horse ready to be put down and then.... 
    \textbf{(e)}~[X] smiling while holding his Colorado State Championship Jersey and wearing all his cycling medals. 
    \textbf{(f)}~Dr. [X] Featured in Ahmedabad Mirror. 
    \textbf{(g)}~[X] (Nepali Youtuber) Lifestyle, Income, Biography, Girlfriend, House, Car. 
    \textbf{(h)}~A boy pauses in front of a wall-sized poster depicting the faces of 90 survivors of the mass killings of Armenians in the Ottoman Empire. 
    \textbf{(i)}~The thighs are the first place of the women's body, where the fat builds up and can be seen easily. For this reason, the women who experienced extra weight are especially unhappy with the appearanc… 
    \textbf{(j)}~China wheelchair specifications lightweight potable comode 24v battery electric wheelchair for disabled people.\\
    The symbol [X] indicates data intentionally anonymized to protect personal information.}
    \label{fig:placeholder}
\end{figure*}

\textbf{First Sample:}
Applying the \textit{Datasheet of Datasets} tool to the first sample of 668~instances from the English subset revealed several technical and compositional issues. Although the initial random sample comprised 1,200 instances, it was reduced to 668 after filtering for valid images using LAION's \textit{img2dataset} library~\cite{beaumont-2021-img2dataset}. Among these, 12 instances returned contextually invalid visual data (e.g., blank white pixels or access-restriction placeholders), while four contained truncated incomplete text captions ending in ellipses (Figs~\ref{fig:horse} and~\ref{fig:weight}).

Regarding privacy and the potential for individual identification, the analysis highlighted significant ethical concerns about exposing minors and sensitive personal information. The sample contained 40 instances pairing personal imagery with explicit sensitive attributes, such as names, ages, specific locations, political opinions, union memberships, and personal relationships, enabling direct identification  (Figs~\ref{fig:text}, ~\ref{fig:mother}, \ref{fig:children1} and \ref{fig:nationality}). Furthermore, similar to \textit{Framed Multi30K}, 18~images depicted children, 11 of which clearly exposed their faces, raising significant privacy and consent concerns for downstream uses (Figs~\ref{fig:mother} and~\ref{fig:children1}).

Explicit  subpopulation identification was sparse and rarely targeted individuals. For instance, nationality was explicitly attributed to a specific person in only one single instance (Fig.~\ref{fig:nationality}), whereas all other occurrences referring to objects or events. Similarly, gender representations were similarly constrained: terms such as ``woman'' or ``women'' appeared in 13 captions, with corresponding human depictions in only 6 images. Likewise, ``man'' or ``men'' occurred in 6 captions, depicting individuals in only 2 images.

Despite automated filtering via \textit{p\_unsafe}, the sample retained 22 instances containing potentially offensive, stereotyping, or anxiety-inducing content, reflecting systematic web-crawl biases. The analysis revealed graphic and distressing subjects, such as a car wreck, houses destroyed by fire, and a horse described as about to be euthanized  (Figs~\ref{fig:house} and \ref{fig:horse}), alongside depictions of historical trauma and geopolitical conflict, including portraits of Armenian genocide survivors (Fig.~\ref{fig:portraits}) and news of police raids on undocumented immigrants. 

Qualitative issues concerning social biases and harmful stereotyping were also prevalent. Specific instances featured ableist language (Fig.~\ref{fig:chair}), body shaming, reinforcement of societal beauty standards (Fig.~\ref{fig:weight}), and subjective, potentially false captions, such as assuming a woman is flirting with an older man (Fig.~\ref{fig:flirting}).

\vspace{0.2cm}
\noindent\textbf{Second Sample:}
Evaluation of the second sample (691 instances) yielded findings highly consistent with the first sample (668 instances) across technical, privacy, and qualitative dimensions.

\begin{figure*}[tbp]
    \centering
    \subfloat[]{\includegraphics[height=2cm,clip,trim={8cm 0 14cm 0}]{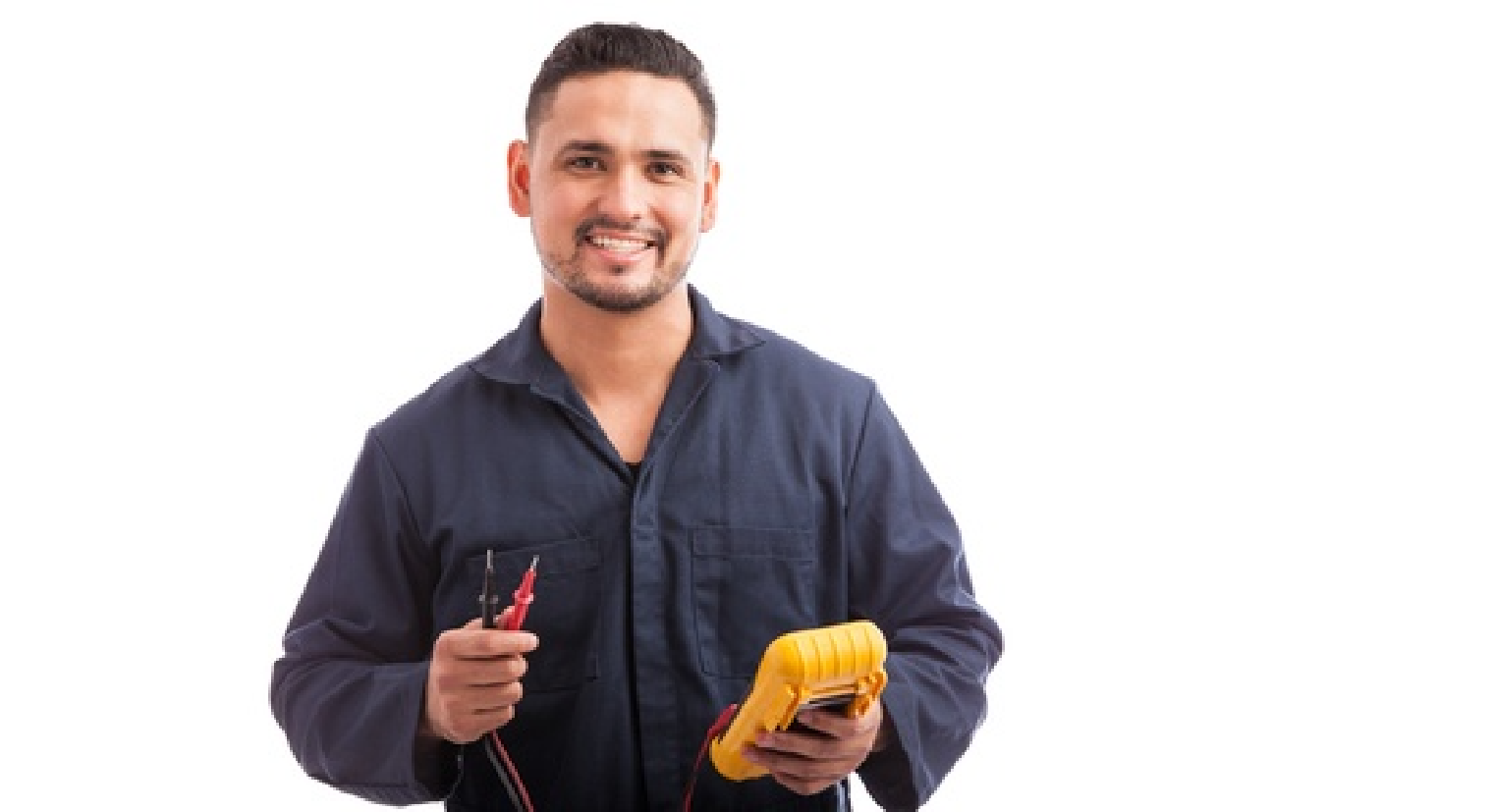}\label{fig:hispanic}}
    \subfloat[]{\includegraphics[height=2cm,clip,trim={3cm 0 3cm 0}]{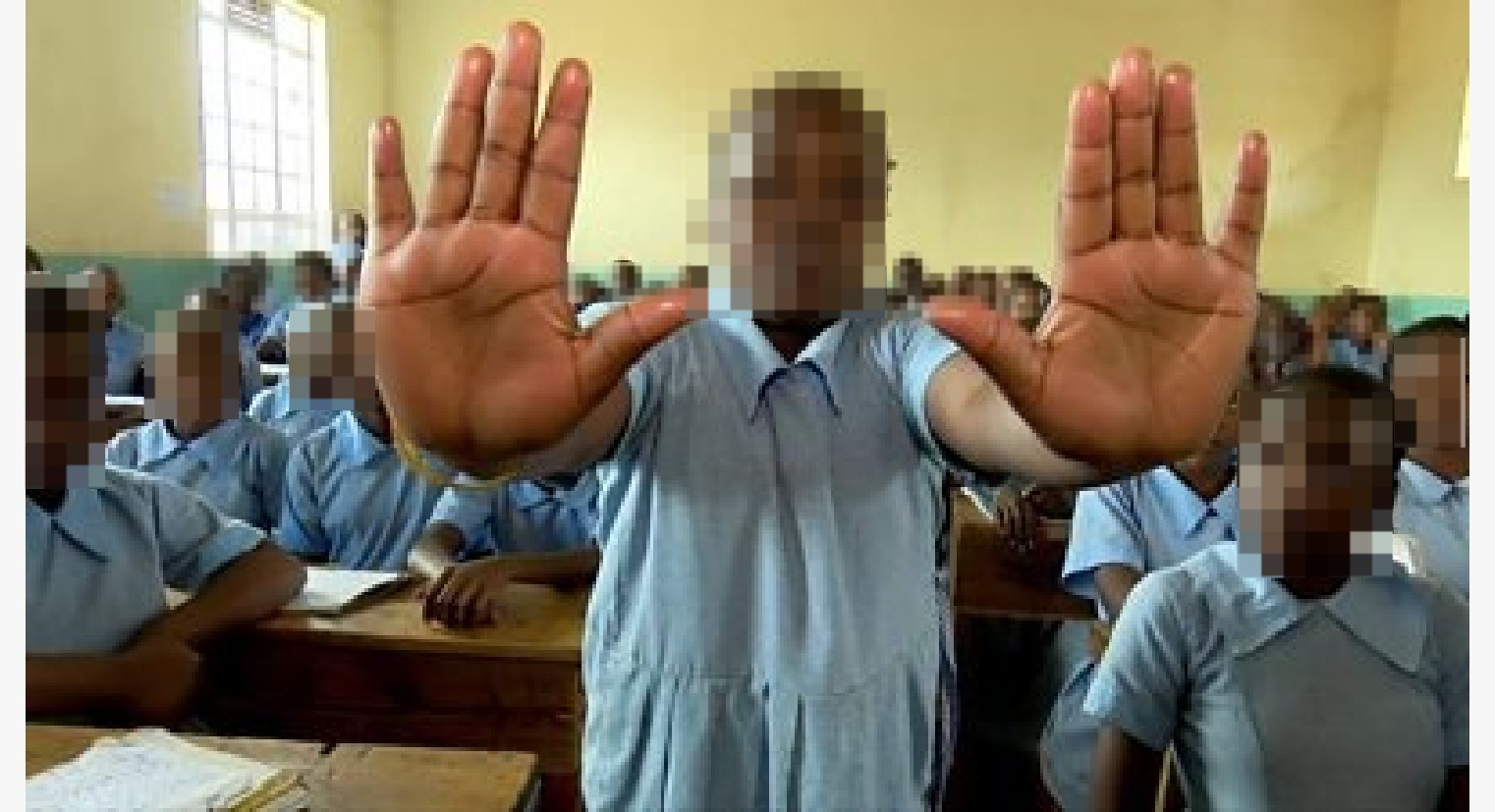}\label{fig:children}}
    \subfloat[]{\includegraphics[height=2cm,clip,trim={9cm 0 9cm 0}]{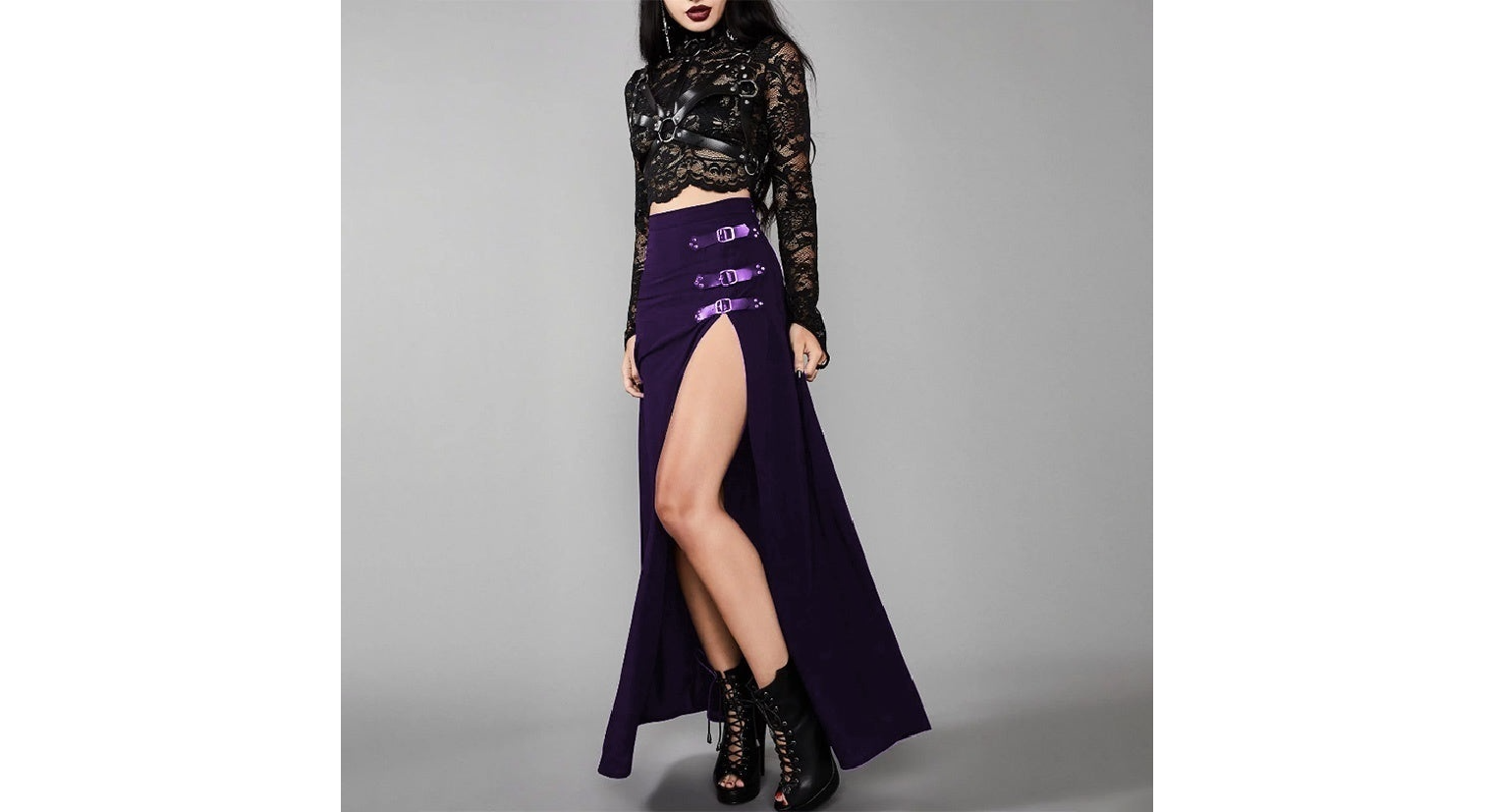}\label{fig:plussize}}
    \subfloat[]{\includegraphics[height=2cm,clip,trim={6cm 0 5cm 0}]{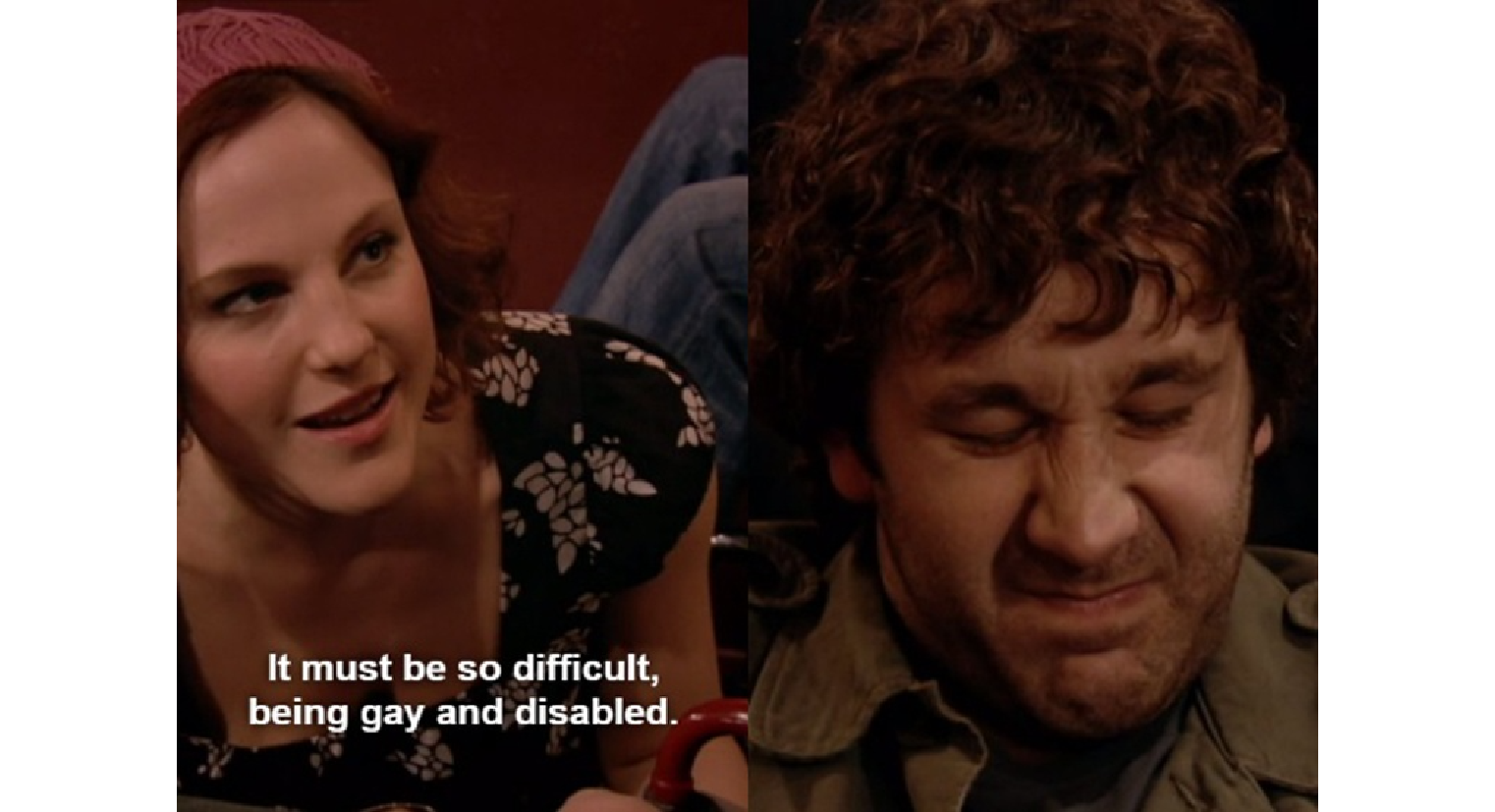}\label{fig:gay}} 
    \subfloat[]{\includegraphics[height=2cm,clip,trim={5cm 0 6cm 0}]{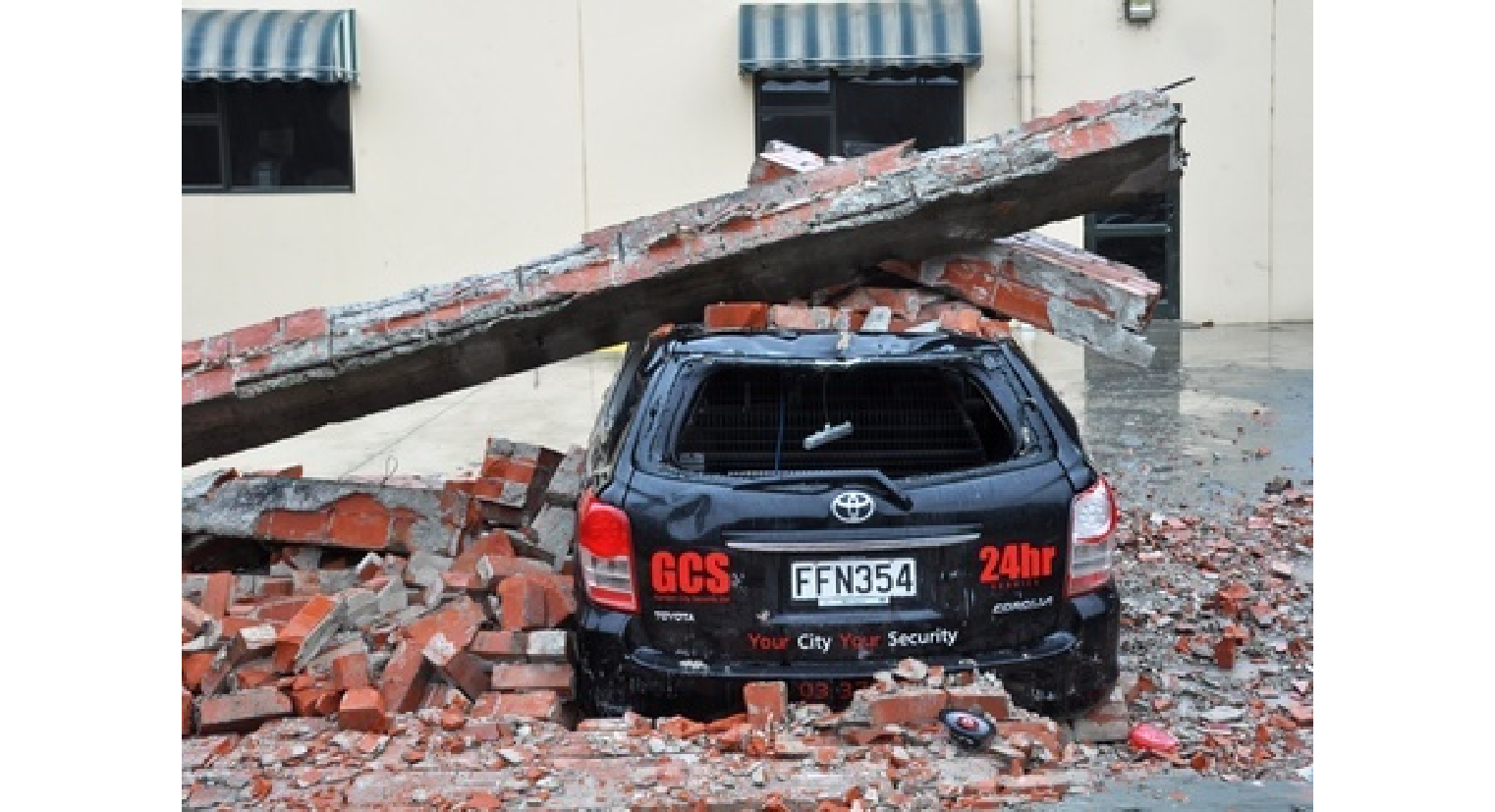}\label{fig:car}}\vspace{0.1cm}\\
    \subfloat[]{\includegraphics[height=2cm,clip,trim={4cm 0 8cm 0}]{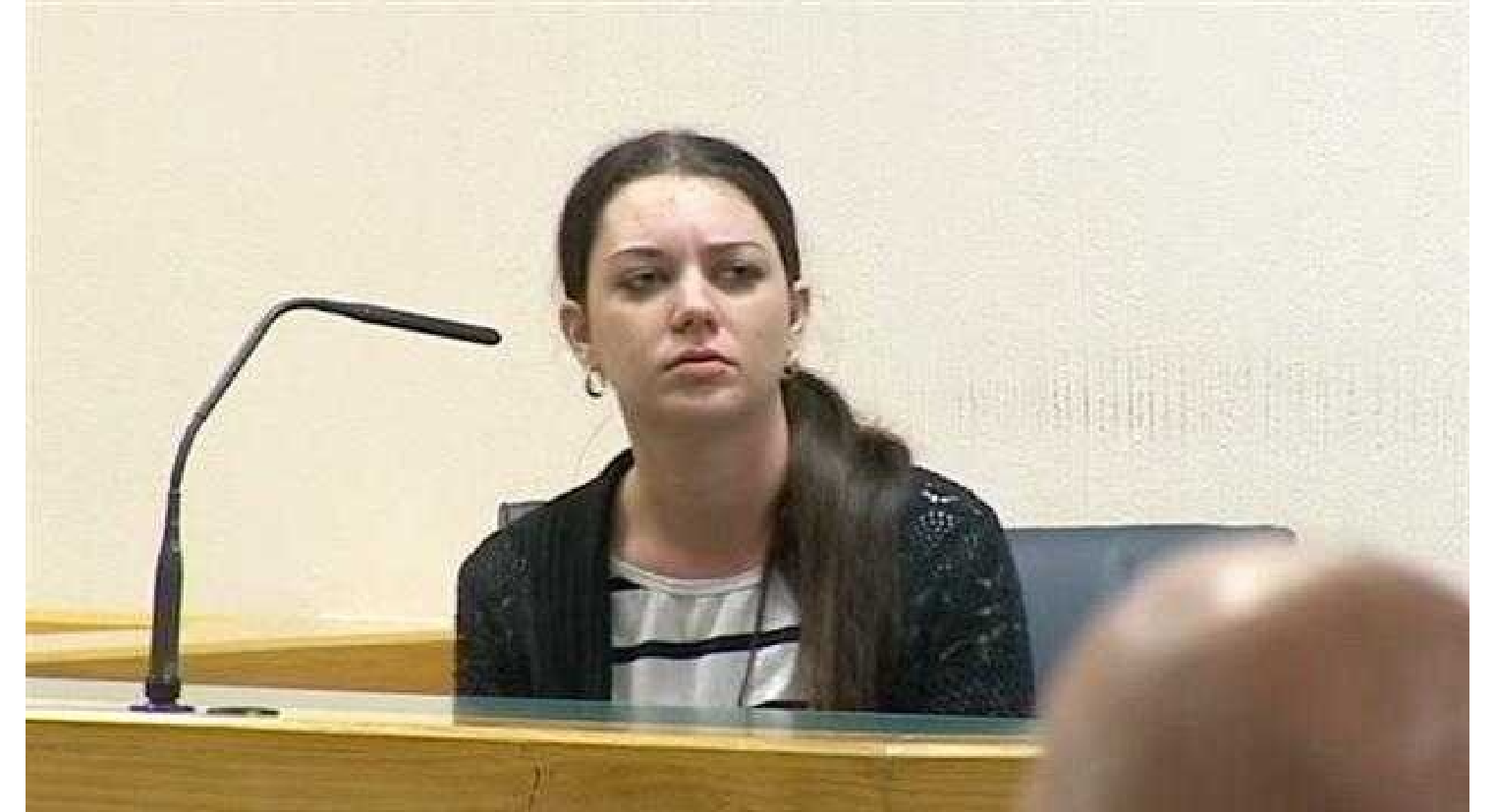}\label{fig:pill}}
    \subfloat[]{\includegraphics[height=2cm,clip,trim={10cm 0 14cm 0}]{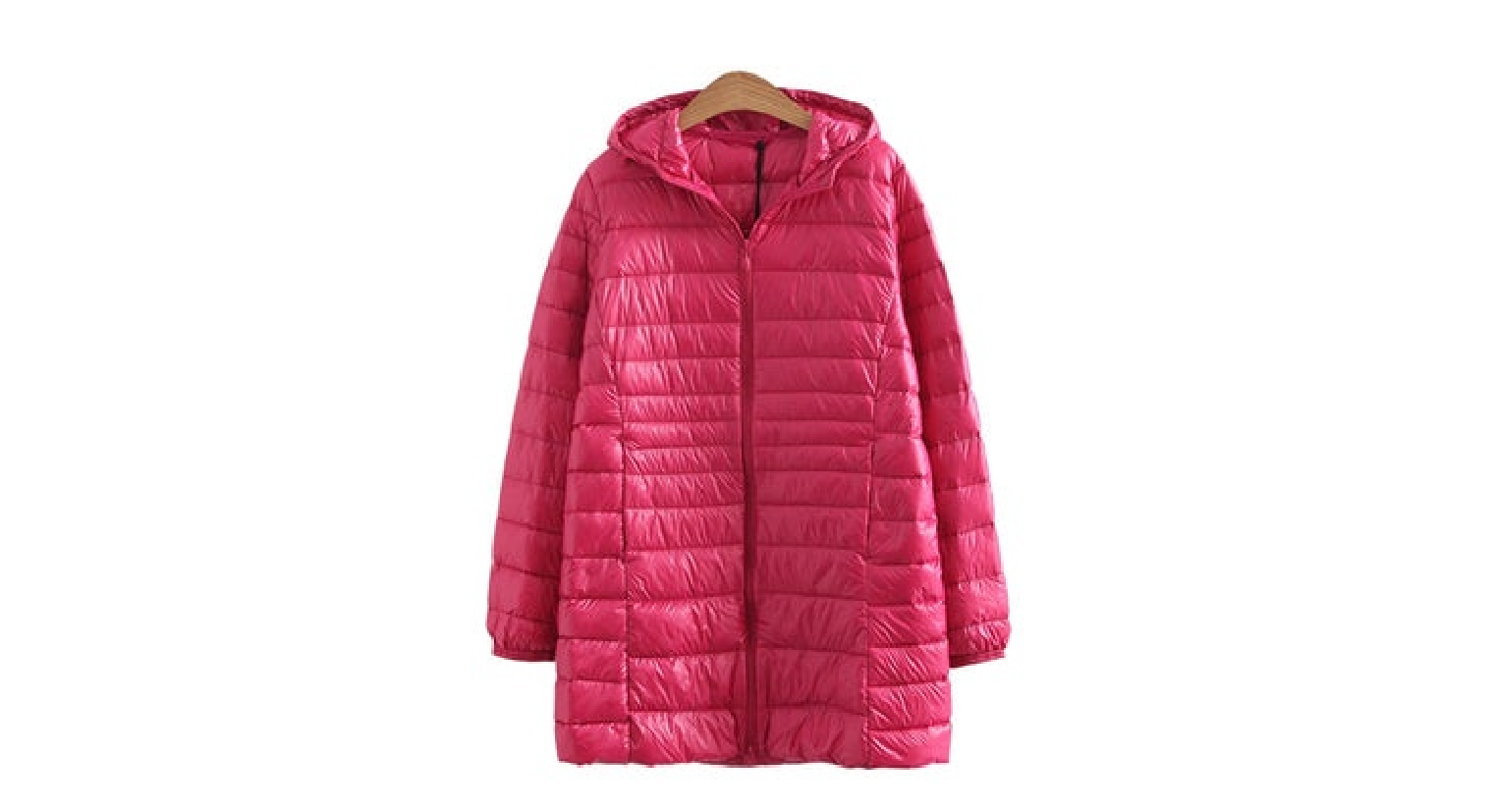}\label{fig:coat}}
    \subfloat[]{\includegraphics[height=2cm,clip,trim={14cm 0 14cm 0}]{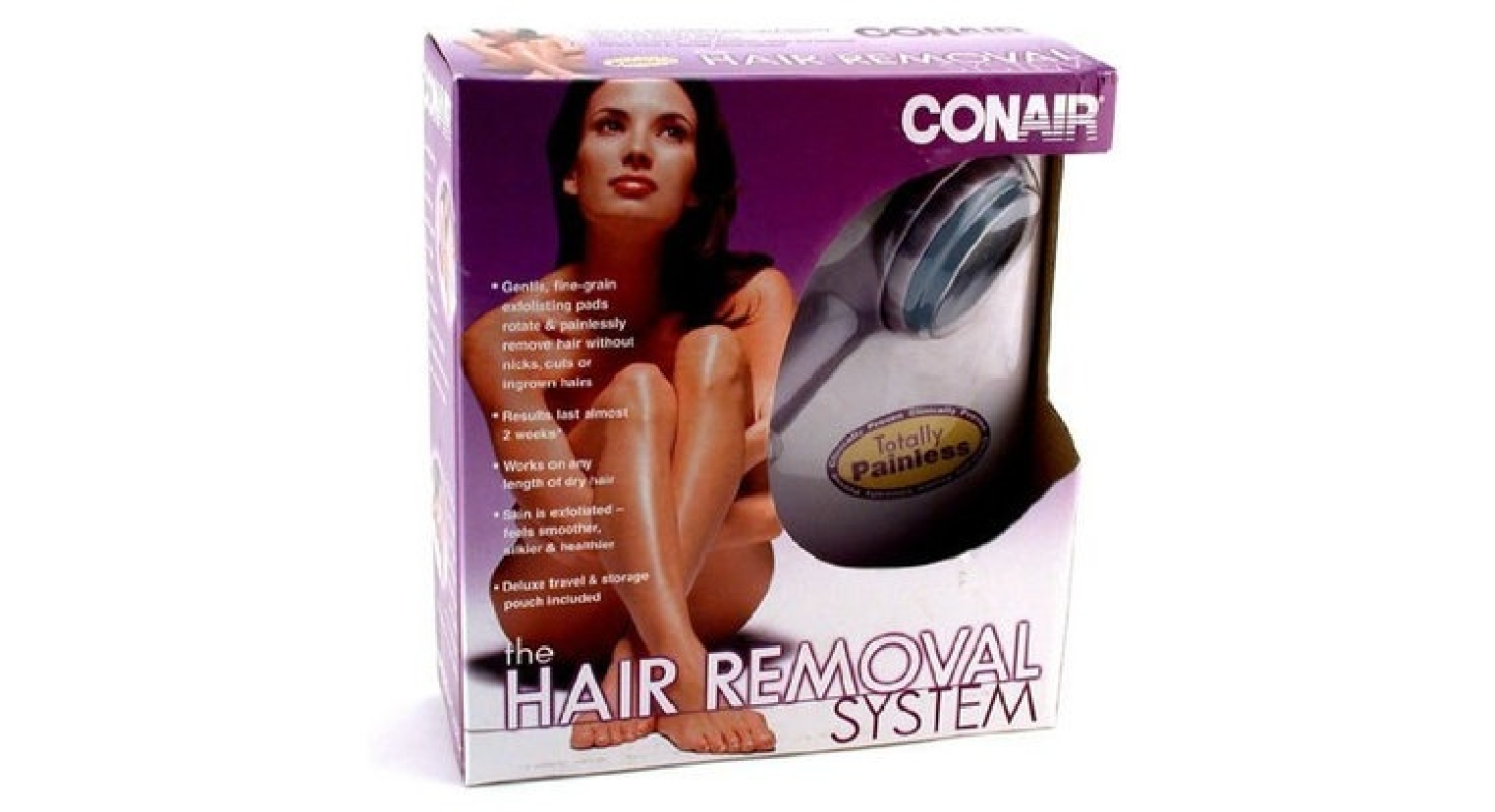}\label{fig:unsafe1}}
    \subfloat[]{\includegraphics[height=2cm,clip,trim={12cm 0 12cm 0}]{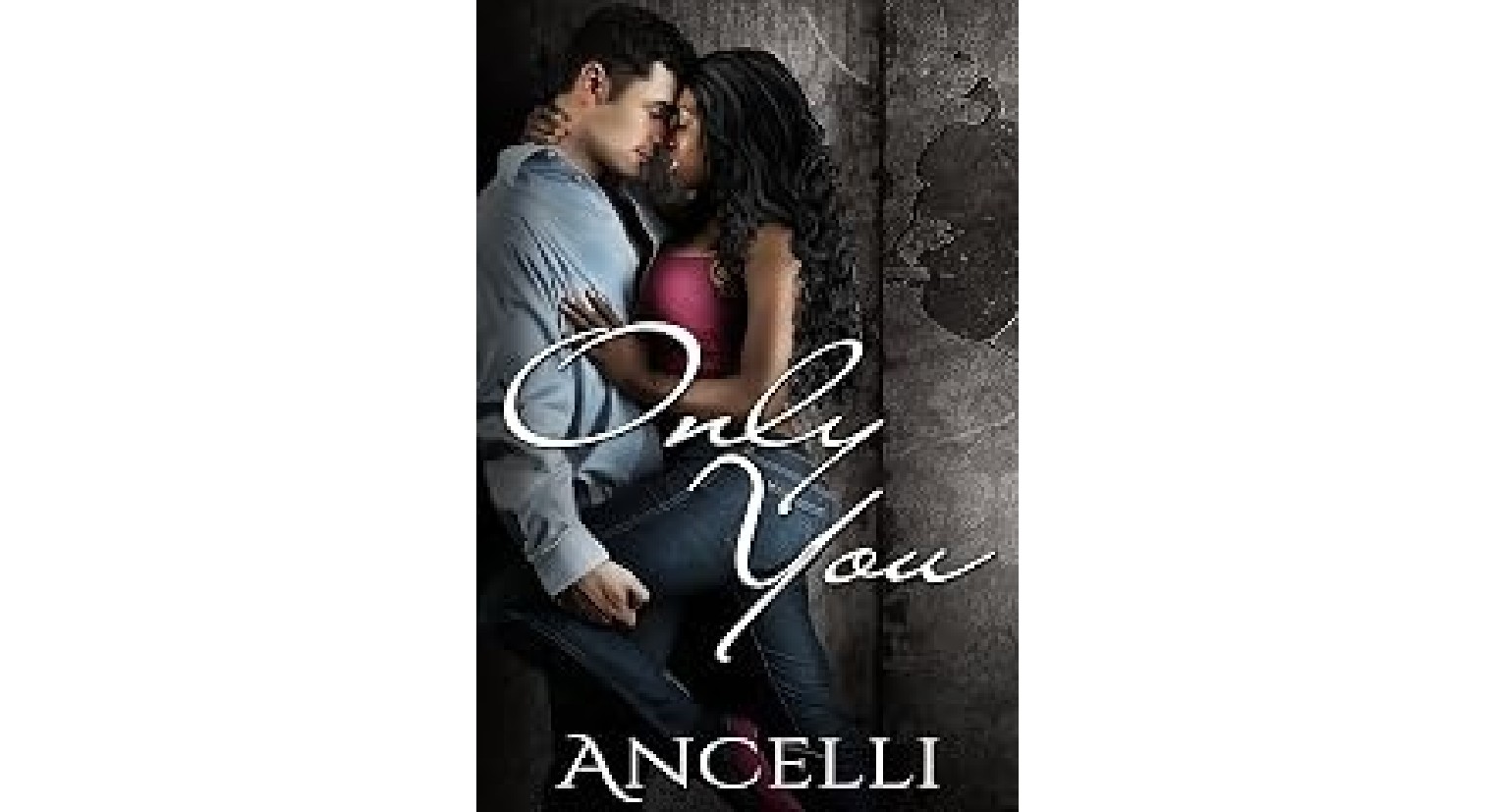}\label{fig:unsafe2}}
    \subfloat[]{\includegraphics[height=2cm,clip,trim={5cm 0 5cm 0}]{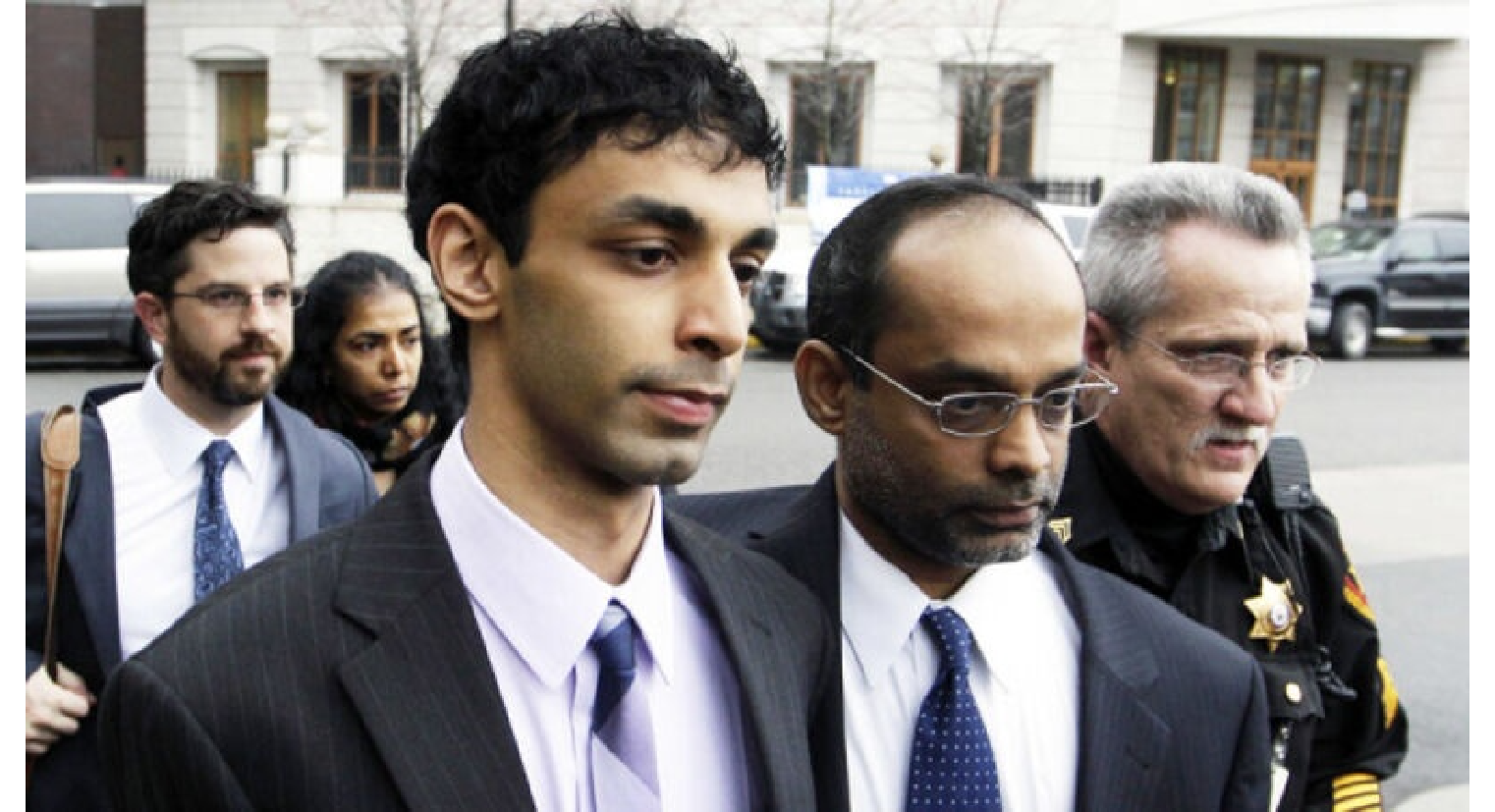}\label{fig:jail}}
    \subfloat[]{\includegraphics[height=2cm,clip,trim={9cm 0 9cm 0}]{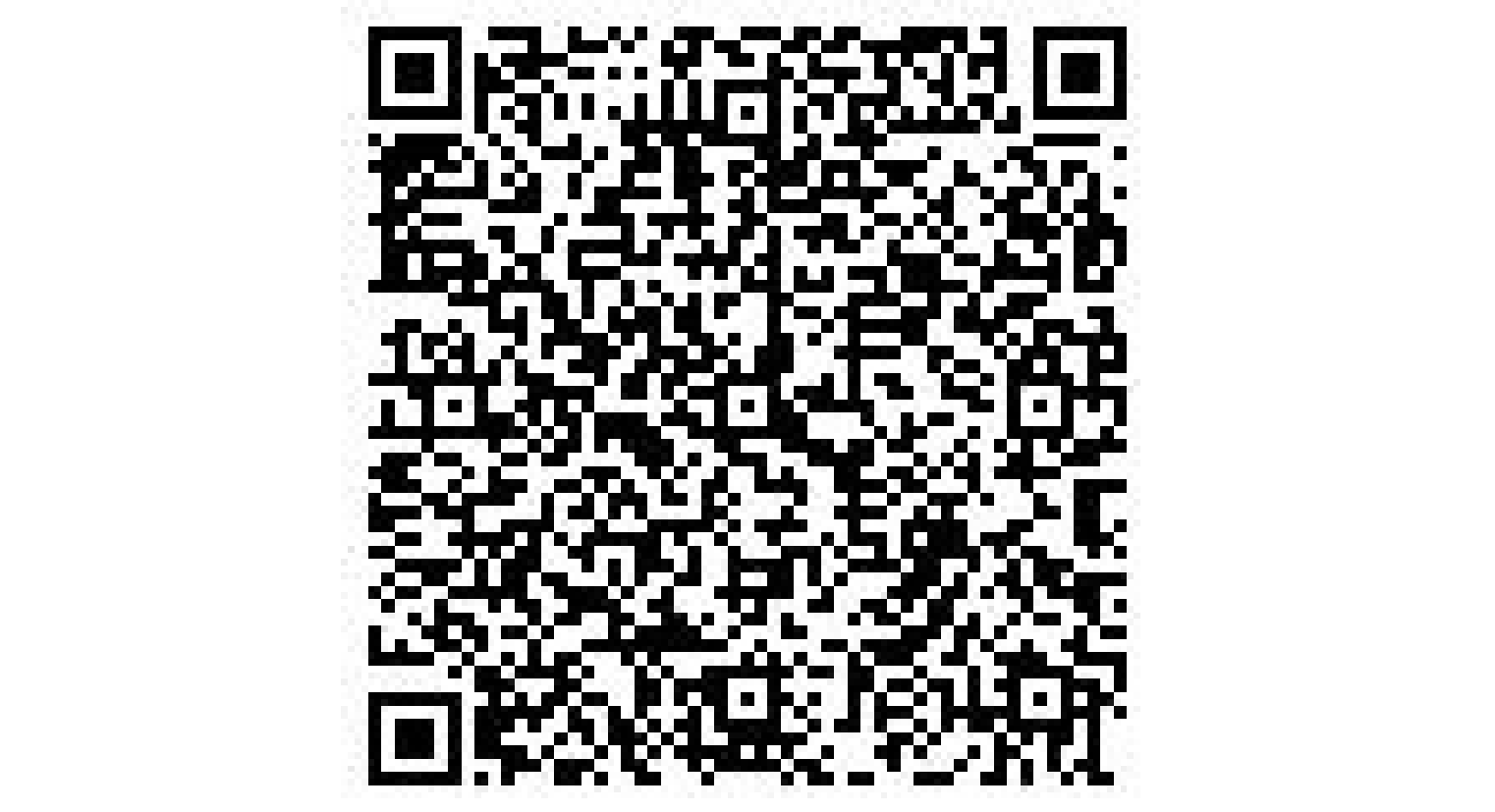}\label{fig:qrcode}}

    \caption{\textbf{Sample instances from \textit{relaion2B-en-research-safe}.} 
    \textbf{(a)}~Portrait of a young Hispanic electrician wearing overalls using a multimeter and smiling on a white background Banque d'images. 
    \textbf{(b)}~Why Kenyan school children are being taught to say 'no ... 
    \textbf{(c)}~Plus Size S-4XL Women Medieval Cosplay Skirts Female Court-Style Gothic High Split Patchwork Long Skirts Sexy High Split Buckle Skirt(no Tops). 
    \textbf{(d)}~The IT Crowd. 
    \textbf{(e)}~Christchurch, New Zealand - March 2011: A car is crushed by a collapsed brick wall on March 20, 2011 in Christchurch following a 6.2 magnitude earthquake. 
    \textbf{(f)}~``[X] testifies about her prescription pill addiction in her husband's Craigslist `companionship' trial.'' 
    \textbf{(g)}~IEMUH Winter Plus Size 90\% White Duck Down Women Jackets Down Warm Coat Female Long Jacket Fat Women Parkas Loose Snow Outwear. 
    \textbf{(h)}~Conair HB1 The Hair Removal System (p\_unsafe 0,005). 
    \textbf{(i)}~Only You Kindle Edition (p\_unsafe 0,005). 
    \textbf{(j)}~[X], Webcam Peep, Gets 30-Day Jail Sentence [Update: Prosecutors Appeal]. 
    \textbf{(k)}~QR code with Turnagain Church Of Christ contacts.\\ 
    The symbol [X] indicates data intentionally anonymized to protect personal information.}
    \label{fig:placeholder2}
\end{figure*}

Filtering the 1,200-instance sample via LAION's \textit{img2dataset} library~\cite{beaumont-2021-img2dataset} reduced the data to 691 image-text pairs. Among these, 9 instances returned invalid visual data (e.g., blank white pixels or access-restriction placeholders). Additionally, 3 instances contained  incomplete text captions terminating in ellipses (Fig.~\ref{fig:children}).

Privacy analysis corroborated severe ethical concerns about minor exposure and sensitive personal information. The sample contained 43 instances pairing personal imagery with explicit sensitive attributes go beyond basic identification, openly revealing details such as individuals' full names, specific locations, religious beliefs, active cellphone numbers, and even sensitive health data and legal history (Figs~\ref{fig:pill}, \ref{fig:jail}, and \ref{fig:qrcode}).  
Additionally, 11 images depicted children, seven of which clearly displayed recognizable facial features (Fig.~\ref{fig:children}).

Subpopulations identification remained sparse and inconsistent. Gender-related terminology occurred in 15 captions for ``woman''/ ``women'' (with corresponding human depictions in 10 images) and 8 captions for ``man''/``men'' (depicting individuals in 3 images). Nationality references primarily denoted objects or events, though several instances contained subjective and potentially inaccurate inferences concerning depicted individuals (Fig.~\ref{fig:hispanic}).

Content safety analysis further highlighted the limitations of the automated \textit{p\_unsafe}. Despite filtering, the sample retained 24 instances of potentially offensive, stereotyping, or anxiety-inducing material. The analysis revealed graphic and distressing subjects, such as car accidents (Fig.~\ref{fig:car}), house fires, and images of violence. Furthermore, specific instances exhibit \textit{p\_unsafe} probabilities that were demonstrably underestimated, failing to flag suggestive intimate acts, images of lingerie, and implicit nudity (Figs~\ref{fig:unsafe1} and~\ref{fig:unsafe2}). 

Qualitative evaluation showed pervasive social biases and harmful stereotyping. Specific instances exhibited ableist and homophobic language, such as an image asserting the difficulty of being both gay and disabled (Fig.~\ref{fig:gay}), toxic social comparisons, and body shaming, including subjective commentary on someone's weight gain.  The dataset also reinforced distorted societal beauty standards by categorizing conventionally thin model as ``plus size'', as well as captions associating ``plus size'' with ``fat women'' (Figs \ref{fig:plussize} and~\ref{fig:coat}). 

In summary, across both \textit{relaion2B-en-research-safe} audited samples, nearly half of the images were broken, blank, or invalid. Furthermore, individual privacy was a major concern; numerous instances exposed sensitive and private information, including full names, active phone numbers, health data and specific locations, alongside identifiable images. Identifiable children's faces further aggravated these consent and privacy concerns. The dataset also contains widespread examples of social biases, harmful stereotyping, and offensive or anxiety-inducing material, underscoring the limitation of relying solely on automated, image-based safety probabilities. 
Ultimately, these empirical findings demonstrate that despite rigorous safety revisions to remove CSAM and NSFW, the dataset remains a snapshot of the internet, heavily embedding societal bias that risk propagating harmful outcomes if models are trained and deployed without rigorous curation.\vspace{0.2cm}

While this study focuses on two multimodal text-image datasets, our  methodology can be applied to other formats, such as video, audio, sensor, and tabular data. Biases and ethical concerns can emerge throughout the data lifecycle, from annotator and subject demographics and consent to data content, collection, curation, and intended applications.

\section{Limitations}
\label{sec:limitations}
Although our methodology can be applied to different datasets, manual auditing data requires data with semantic content that can be meaningfully interpreted by human auditors, limiting its applicability to certain types of datasets.

This study is limited to four general-purpose AIETs, selected to support the applicability of our methodology across different types of data. Other tools, particularly those designed for specific purposes or dataset types, may reveal additional concerns and offer different perspectives and approaches to dataset auditing.

Analyzing a small sample does not make it possible to identify every bias and ethical problem in a dataset, nor would analyzing the entire dataset, given the human limitations involved. This is because ethical and critical thinking depends on situated knowledge, different repertoires, critical opinions, and contexts of the individuals involved. Thus, the analysis and its results may vary depending on who conducts it. Since this research was conducted by only computer science and linguistics researchers and our audit is restricted by our geographic and cultural context, systemic harms, slurs, and political sensitivities affecting other marginalized populations may have overlooked. Therefore, the biases and ethical problems identified in this paper do not necessarily reflect all those present in the sample. The work would benefit from researchers with diverse academic backgrounds, locations, and social contexts. 

This study uses very small samples due to time and personnel constraints, which may prevent critical biases and problems from emerging. Although our method aims to identify biases in dataset creation, small random samples cannot reliably estimate the frequency or prevalence of concerns across the entire dataset, and rare but critical problems may go unnoticed. Larger samples could help address this limitation. Furthermore, the audit, tagging, and analysis processes are subjective, and their reliability was not formally assessed. Different groups may therefore categorize the same samples differently, affecting the reproducibility of our results.

\section{Conclusion}
\label{sec:conclusion}
Applying AIETs to such small samples of the datasets demonstrated that conducting ethical audits of large datasets using small samples is viable. This methodology challenges the claim that datasets are ``too big to inspect'', offering a concrete and accessible starting point for ethical evaluations that might otherwise seem unfeasible given the scale and complexity of the data involved. By reducing the need for extensive computational resources, this method democratizes dataset audits and allows the community to demand manual and qualitative audits before datasets are published.

During the evaluation, the tools' questions provoked different perspectives and approaches to analyzing the same datasets. This process allowed us to identify, for example, in the \textit{Framed Multi30K} dataset, linguistic and cultural biases present in how the data identifies subpopulations, pointing to a tendency to mark or annotate only ethnicities and origins considered non-standard, which reinforces the concept of ``white'' as the non-racialized default. 
Similarly, in the \textit{relaion2B-en-research-safe} dataset, we found that automated safety classifications (\textit{p\_unsafe} tag) based only on images fail to capture multimodal and contextual harms. Thus, the dataset remains a snapshot of the internet, heavily embedding societal bias that can create harmful outcomes if models are trained and deployed without further curation.

However, the application also revealed that AIETs alone are not sufficient to  expose biases. Although they foster new perspectives, the ethical analysis relies fundamentally on the human factor, guided by the diverse knowledge, backgrounds, and perspectives of the research group. 
Thus, dataset auditing must be operate  as an iterative, continuous process involving multidisciplinary teams throughout the dataset's lifecycle. Moreover, AIETs and other forms of documentation should become standard practice, promoting transparency, defining usage boundaries, and warning stakeholders of potential harms.

\section{Generative AI Usage Statement}
To prepare this manuscript, we used the \textit{Gemini 3.5 Flash} model to translate text from Brazilian Portuguese summaries and reports on this research into  English and formatted images in \LaTeX. The tool served only as technical support, with the intellectual content and integrity of the article remaining under the responsibility of the authors.

\section{Ethical Considerations Statements}
Computer science and linguistics researchers conducted this research, and  biases from these areas may affect the interpretation and application of the AIETs and their results. Bias auditing is not a neutral process~\cite{baeza2018bias} and depends on the repertoire, previous experiences, and perception of those involved.

This work uses samples from public datasets. The authors acknowledge the presence of sensitive data, such as images of minors, names, locations, and other private information, which were kept solely within the group for academic purposes to avoid privacy and security risks.

This work does not aim to justify or identify all biases within \textit{Framed Multi30K} and \textit{relaion2B-en-research-safe}, which would be unfeasible. We show that, in automated uncurated datasets and even in well-structured,  carefully curated datasets, a small random sample can reveal biases, representation problems, and technical inconsistencies. 

In the case of \textit{Framed Multi30k}, part of these inconsistencies and biases derive from the original \textit{Flickr30K} datasets. The decision to keep these cases in the discussion of this work was made to enable the study and debate around divergences in annotations, both automatic and made directly by humans, whose sociodemographic aspects and perspectives on ethnicities and gender by \textit{Flickr30K} annotators and by those of \textit{Framed Multi30K} are documented in~\mbox{\citet{Viridiano2024FM30K}.}

However, even with an understanding of the origin and intent of these processes, such biases remain part of the dataset as a whole, demanding ongoing discussion and analysis. Thus, this study proposes a practical and effective method for auditing large datasets to prevent the perpetuation and propagation of biases.

Finally, this work does not aim to discourage the use of \textit{Framed Multi30K} or \textit{relaion2B-en-research-safe}. Instead, this ethical analysis aims to offer transparency, support documentation, and enable proper, conscious use of these datasets.

\section{Author Contributions}
G.~Marchetti conducted the study under the supervision of C.~Caetano and S.~Avila, and wrote this manuscript with support from J.~Silva. T.~Torrent assisted G.~Marchetti understand the \textit{Framed Multi30k} dataset and complete the \textit{Datasheets for Datasets}, \textit{Data Statements}, \textit{Data Nutrition Labels}, and \textit{Data Cards}. All authors reviewed and approved the final version of this manuscript and agree to be accountable for all aspects of the work, ensuring its accuracy and integrity.

\section*{Acknowledgments} 
This project was supported by the Ministry of Science, Technology, and Innovation of Brazil, with resources granted by the Federal Law 8.248 of October 23, 1991, under the PPI-Softex. The project was coordinated by Softex and published as Intelligent agents for mobile platforms based on Cognitive Architecture technology [01245.003479/2024-10]. J.~Silva is funded by FAPESP 2024/23118-1 and was partially financed by the Coordination for the Improvement of Higher Education Personnel (CAPES) -- Finance Code 001.
T.~Torrent is also partially funded by CNPq (311241/2025-5). His participation in this research relies on funding granted to INCT-TILDIAR (CNPq grant 408490/2024-1). S.~Avila is also partially funded by FAPESP 2023/12086-9, 2023/12865-8, \mbox{2020/09838-0}, 2013/08293-7, and CNPq 316489/2023-9.

\bibliographystyle{ACM-Reference-Format}
\bibliography{sample-base}

@String{Computing = "Computing" }

@String{Computer = "{IEEE} Computer" }

@String{Chelsea = "Chelsea" }

@article{gebru2021datasheets,
    author = {Gebru, Timnit and Morgenstern, Jamie and Vecchione, Briana and Vaughan, Jennifer Wortman and Wallach, Hanna and III, Hal Daum\'{e} and Crawford, Kate},
    title = {Datasheets for datasets},
    year = {2021},
    volume = {64},
    number = {12},
    journal = {Commun. ACM},
    pages = {86–92},
    numpages = {7}
}

@article{10.1145/3594737,
author = {McMillan-Major, Angelina and Bender, Emily M. and Friedman, Batya},
title = {Data Statements: From Technical Concept to Community Practice},
year = {2024},
volume = {1},
number = {1},
journal = {ACM J. Responsib. Comput.},
articleno = {1},
numpages = {17},
}

@article{chmielinski2022dataset,
  title={The dataset nutrition label (2nd gen): Leveraging context to mitigate harms in artificial intelligence},
  author={Chmielinski, Kasia S and Newman, Sarah and Taylor, Matt and Joseph, Josh and Thomas, Kemi and Yurkofsky, Jessica and Qiu, Yue Chelsea},
  journal={arXiv preprint arXiv:2201.03954},
  year={2022}
}

@article{bender2018data,
  title={Data statements for natural language processing: Toward mitigating system bias and enabling better science},
  author={Bender, Emily M and Friedman, Batya},
  journal={Transactions of the Association for Computational Linguistics},
  volume={6},
  pages={587--604},
  year={2018},
  publisher={MIT Press One Rogers Street, Cambridge, MA 02142-1209, USA journals-info~…}
}

@misc{holland2018,
  title = {The {{Dataset Nutrition Label}}: {{A Framework To Drive Higher Data Quality Standards}}},
  shorttitle = {The {{Dataset Nutrition Label}}},
  author = {Holland, Sarah and Hosny, Ahmed and Newman, Sarah and Joseph, Joshua and Chmielinski, Kasia},
  year = {2018},
  number = {arXiv:1805.03677},
  eprint = {1805.03677},
  publisher = {arXiv},
  archiveprefix = {arXiv}
}

@inproceedings{viridiano2024framed,
  title={{Framed Multi30K:} A frame-based multimodal-multilingual dataset},
  author={Viridiano, Marcelo and Lorenzi, Arthur and Torrent, Tiago Timponi and Matos, Ely E and Pagano, Adriana S and Sigiliano, Nat{\'a}lia Sathler and Gamonal, Maucha and de Andrade Abreu, Helen and others},
  booktitle={Joint International Conference on Computational Linguistics, Language Resources and Evaluation},
  pages={7438--7449},
  year={2024}
}

@inproceedings{elliott2016multi30k,
  title={Multi30k: Multilingual english-german image descriptions},
  author={Elliott, Desmond and Frank, Stella and Sima’an, Khalil and Specia, Lucia},
  booktitle={Workshop on Vision and Language},
  pages={70--74},
  year={2016}
}

@inproceedings{plummer2015flickr30k,
  title={Flickr30k entities: Collecting region-to-phrase correspondences for richer image-to-sentence models},
  author={Plummer, Bryan A and Wang, Liwei and Cervantes, Chris M and Caicedo, Juan C and Hockenmaier, Julia and Lazebnik, Svetlana},
  booktitle={IEEE International Conference on Computer Vision},
  pages={2641--2649},
  year={2015}
}

@inproceedings{xia2021lome,
  title={LOME: Large ontology multilingual extraction},
  author={Xia, Patrick and Qin, Guanghui and Vashishtha, Siddharth and Chen, Yunmo and Chen, Tongfei and May, Chandler and Harman, Craig and Rawlins, Kyle and White, Aaron Steven and Van Durme, Benjamin},
  booktitle={Conference of the European Chapter of the Association for Computational Linguistics: System Demonstrations},
  pages={149--159},
  year={2021}
}

@article{silva2026evaluation,
  title={Evaluation of AI ethics tools in language models: a developers’ perspective case study},
  author={Silva, Jhessica and Moreira, Diego and O dos Santos, Gabriel and Ferreira, Alef and Maia, Helena and Avila, Sandra and Pedrini, Helio},
  journal={AI and Ethics},
  volume={6},
  number={2},
  pages={210},
  year={2026},
}

@article{thiel2023identifying,
  title={Identifying and eliminating csam in generative ml training data and models},
  author={Thiel, David},
  journal={Stanford Internet Observatory, Cyber Policy Center, December},
  volume={23},
  number={3},
  pages={131},
  year={2023}
}

@article{birhane2021multimodal,
  title={Multimodal datasets: misogyny, pornography, and malignant stereotypes},
  author={Birhane, Abeba and Prabhu, Vinay Uday and Kahembwe, Emmanuel},
  journal={arXiv preprint arXiv:2110.01963},
  year={2021}
}

@article{dominguez2026unmasking,
  title={Unmasking LAION-5B: Age, Gender, Race, and Emotion Biases in Large-Scale Image Datasets},
  author={Dominguez-Catena, Iris and Paternain, Daniel and Galar, Mikel},
  journal={arXiv preprint arXiv:2606.23204},
  year={2026}
}

@article{schuhmann2022laion,
  title={Laion-5b: An open large-scale dataset for training next generation image-text models},
  author={Schuhmann, Christoph and Beaumont, Romain and Vencu, Richard and Gordon, Cade and Wightman, Ross and Cherti, Mehdi and Coombes, Theo and Katta, Aarush and Mullis, Clayton and Wortsman, Mitchell and others},
  journal={Advances in neural information processing systems},
  volume={35},
  pages={25278--25294},
  year={2022}
}

@article{birhane2023hate,
  title={On hate scaling laws for data-swamps},
  author={Birhane, Abeba and Prabhu, Vinay and Han, Sang and Boddeti, Vishnu Naresh},
  journal={arXiv preprint arXiv:2306.13141},
  year={2023}
}

@inproceedings{bender2021dangers,
  title={On the dangers of stochastic parrots: Can language models be too big?},
  author={Bender, Emily M and Gebru, Timnit and McMillan-Major, Angelina and Shmitchell, Shmargaret},
  booktitle={ACM Conference on Fairness, Accountability, and Transparency},
  pages={610--623},
  year={2021}
}

@Book{BPLN_livro_4ed:2026,
   title     = {Processamento de Linguagem Natural: Conceitos,
                Técnicas e Aplicações em Português},
   editor    = {Caseli, H. M. and Nunes, M. G. V.},
   year      = {2026},
   publisher = {BPLN},   
   edition   = {4},
   isbn      = {978-65-02-00178-3},
   url       = {https://brasileiraspln.ufscar.br/livro-pln-4ed}
}

@article{young-etal-2014-image,
    title = "From image descriptions to visual denotations: New similarity metrics for semantic inference over event descriptions",
    author = "Young, Peter  and
      Lai, Alice  and
      Hodosh, Micah  and
      Hockenmaier, Julia",
    journal = "Transactions of the Association for Computational Linguistics",
    volume = "2",
    year = "2014",
    address = "Cambridge, MA",
    publisher = "MIT Press",
    url = "https://aclanthology.org/Q14-1006",
    doi = "10.1162/tacl_a_00166",
    pages = "67--78",
}

@article{10.3389/fpsyg.2022.838441,
  author={Torrent, Tiago Timponi and Matos, Ely Edison da Silva and Belcavello, Frederico and Viridiano, Marcelo and Gamonal, Maucha Andrade and Costa, Alexandre Diniz da and Marim, Mateus Coutinho},   
  title={Representing Context in FrameNet: A Multidimensional, Multimodal Approach},      
  journal={Frontiers in Psychology},      
  volume={13},      
  year={2022},      
}

@incollection{framenet2009,
	address = {Oxford, {UK}},
	title = {A frames approach to semantic analysis},
	url = {http://lingo.stanford.edu/sag/papers/Fillmore-Baker-2011.pdf},
	booktitle = {The Oxford Handbook of Linguistic Analysis},
	publisher = {Oxford University Press},
	author = {Fillmore, Charles J. and Baker, Collin},
	editor = {Heine, Bernd and Narrog, Heiko},
	month = dec,
	year = {2009},
	pages = {791--816}
}

@phdthesis{Viridiano2024FM30K,
  author    = {Marcelo Viridiano},
  title     = {Framed Multi30K: Um dataset multimodal-multilíngue baseado em semântica de frames},
  school    = {Universidade Federal de Juiz de Fora},
  year      = {2024},
  address   = {Juiz de Fora},
  type      = {Tese de Doutorado},
  advisor   = {Tiago Timponi Torrent},
  language  = {Portuguese}
}

@article{baeza2018bias,
  title={Bias on the web},
  author={Baeza-Yates, Ricardo},
  journal={Communications of the ACM},
  volume={61},
  number={6},
  pages={54--61},
  year={2018},
  publisher={ACM New York, NY, USA}
}

@inproceedings{reynolds2026reflection,
  title={From Reflection to Repair: A Scoping Review of Dataset Documentation Tools},
  author={Reynolds-Cu{\'e}llar, Pedro and Wong-Villacres, Marisol and Alvarado Garcia, Adriana and Precel, Heila},
  booktitle={Conference on Human Factors in Computing Systems},
  year={2026}
}

@inproceedings{pushkarna2022data,
  title={Data cards: Purposeful and transparent dataset documentation for responsible ai},
  author={Pushkarna, Mahima and Zaldivar, Andrew and Kjartansson, Oddur},
  booktitle={ACM conference on fairness, accountability, and transparency},
  pages={1776--1826},
  year={2022}
}

@misc{relaion,
  author = {LAION},
  title = {Releasing Re-LAION-5B: transparent iteration on LAION-5B with additional safety fixes},
  howpublished = {\url{https://laion.ai/blog/relaion-5b/}},
  note = {Accessed: 30 aug, 2024},
  year = {2024}
}

@inproceedings{belcavello2022charon,
  title={Charon: A FrameNet annotation tool for multimodal corpora},
  author={Belcavello, Frederico and Viridiano, Marcelo and da Silva Matos, Ely Edison and Torrent, Tiago Timponi},
  booktitle={Linguistic Annotation Workshop (LAW-XVI) within LREC2022},
  pages={91--96},
  year={2022}
}

@article{birhane2023into,
  title={Into the laion’s den: Investigating hate in multimodal datasets},
  author={Birhane, Abeba and Han, Sanghyun and Boddeti, Vishnu and Luccioni, Sasha and others},
  journal={Advances in neural information processing systems},
  volume={36},
  pages={21268--21284},
  year={2023}
}

@misc{dastin2018amazon,
  author       = {Jeffrey Dastin},
  title        = {Amazon scraps secret {AI} recruiting tool that showed bias against women},
  howpublished = {Reuters},
  year         = {2018},
  month        = {oct},
  day          = {10},
  url          = {https://www.reuters.com/article/world/insight-amazon-scraps-secret-ai-recruiting-tool-that-showed-bias-against-women-idUSKCN1MK0AG/}
}

@article{obermeyer2019dissecting,
  title={Dissecting racial bias in an algorithm used to manage the health of populations},
  author={Obermeyer, Ziad and Powers, Brian and Vogeli, Christine and Mullainathan, Sendhil},
  journal={Science},
  volume={366},
  number={6464},
  pages={447--453},
  year={2019},
}

@article{luccioni2023stable,
  title={Stable bias: Evaluating societal representations in diffusion models},
  author={Luccioni, Sasha and Akiki, Christopher and Mitchell, Margaret and Jernite, Yacine},
  journal={Advances in neural information processing systems},
  volume={36},
  pages={56338--56351},
  year={2023}
}

@misc{daneshjou2021disparities,
  author       = {Roxana Daneshjou and Kailas Vodrahalli and Weixin Liang and Roberto A. Novoa and Melissa Jenkins and Veronica Rotemberg and Justin Ko and Susan M. Swetter and Elizabeth E. Bailey and Olivier Gevaert and Pritam Mukherjee and Michelle Phung and Kiana Yekrang and Bradley Fong and Rachna Sahasrabudhe and James Zou and Albert Chiou},
  title        = {Disparities in Dermatology {AI}: Assessments Using Diverse Clinical Images},
  year         = {2021},
  eprint       = {2111.08006},
  archivePrefix= {arXiv},
}

@article{wen2022characteristics,
  author  = {David Wen and Saad M. Khan and Antonio Ji Xu and Hussein Ibrahim and Luke Smith and Jose Caballero and Luis Zepeda and Carlos de Blas Perez and Alastair K. Denniston and Xiaoxuan Liu and Rubeta N. Matin},
  title   = {Characteristics of publicly available skin cancer image datasets: a systematic review},
  journal = {The Lancet Digital Health},
  year    = {2022},
  volume  = {4},
  number  = {1},
  pages   = {e64--e74},
}

@misc{nicoletti2023humans,
  author       = {Leonardo Nicoletti and Dina Bass},
  title        = {Humans Are Biased: Generative {AI} Is Even Worse},
  howpublished = {Bloomberg},
  year         = {2023},
  month        = {jun},
  day          = {23},
  url          = {https://www.bloomberg.com/graphics/2023-generative-ai-bias/}
}

@article{kaplan2020scaling,
  title={Scaling laws for neural language models},
  author={Kaplan, Jared and McCandlish, Sam and Henighan, Tom and Brown, Tom B and Chess, Benjamin and Child, Rewon and Gray, Scott and Radford, Alec and Wu, Jeffrey and Amodei, Dario},
  journal={arXiv preprint arXiv:2001.08361},
  year={2020}
}

@article{hoffmann2022training,
  title={Training compute-optimal large language models},
  author={Hoffmann, Jordan and Borgeaud, Sebastian and Mensch, Arthur and Buchatskaya, Elena and Cai, Trevor and Rutherford, Eliza and Casas, Diego de Las and Hendricks, Lisa Anne and Welbl, Johannes and Clark, Aidan and others},
  journal={arXiv preprint arXiv:2203.15556},
  year={2022}
}

@inproceedings{dodge2021documenting,
  title={Documenting large webtext corpora: A case study on the colossal clean crawled corpus},
  author={Dodge, Jesse and Sap, Maarten and Marasovi{\'c}, Ana and Agnew, William and Ilharco, Gabriel and Groeneveld, Dirk and Mitchell, Margaret and Gardner, Matt},
  booktitle={Conference on Empirical Methods in Natural Language Processing},
  pages={1286--1305},
  year={2021}
}

@inproceedings{luccioni2021whats,
  title={What’s in the box? an analysis of undesirable content in the Common Crawl corpus},
  author={Luccioni, Alexandra and Viviano, Joseph},
  booktitle={Annual Meeting of the Association for Computational Linguistics and the 11th International Joint Conference on Natural Language Processing (Volume 2: Short Papers)},
  pages={182--189},
  year={2021}
}

@article{paullada2021data,
  title={Data and its (dis) contents: A survey of dataset development and use in machine learning research},
  author={Paullada, Amandalynne and Raji, Inioluwa Deborah and Bender, Emily M and Denton, Emily and Hanna, Alex},
  journal={Patterns},
  volume={2},
  number={11},
  year={2021},
  publisher={Elsevier}
}

@article{bandy2021addressing,
  title={Addressing" documentation debt" in machine learning research: A retrospective datasheet for bookcorpus},
  author={Bandy, Jack and Vincent, Nicholas},
  journal={arXiv preprint arXiv:2105.05241},
  year={2021}
}

@inproceedings{sambasivan2021everyone,
  title={“Everyone wants to do the model work, not the data work”: Data Cascades in High-Stakes AI},
  author={Sambasivan, Nithya and Kapania, Shivani and Highfill, Hannah and Akrong, Diana and Paritosh, Praveen and Aroyo, Lora M},
  booktitle={Conference on Human Factors in Computing Systems},
  pages={1--15},
  year={2021}
}

@article{raji2021ai,
  title={AI and the everything in the whole wide world benchmark},
  author={Raji, Inioluwa Deborah and Bender, Emily M and Paullada, Amandalynne and Denton, Emily and Hanna, Alex},
  journal={arXiv preprint arXiv:2111.15366},
  year={2021}
}

@article{longpre2024large,
  title={A large-scale audit of dataset licensing and attribution in AI},
  author={Longpre, Shayne and Mahari, Robert and Chen, Anthony and Obeng-Marnu, Naana and Sileo, Damien and Brannon, William and Muennighoff, Niklas and Khazam, Nathan and Kabbara, Jad and Perisetla, Kartik and others},
  journal={Nature Machine Intelligence},
  volume={6},
  number={8},
  pages={975--987},
  year={2024},
  publisher={Nature Publishing Group UK London}
}

@article{longpre2024consent,
  title={Consent in crisis: The rapid decline of the ai data commons},
  author={Longpre, Shayne and Mahari, Robert and Lee, Ariel and Lund, Campbell and Oderinwale, Hamidah and Brannon, William and Saxena, Nayan and Obeng-Marnu, Naana and South, Tobin and Hunter, Cole and others},
  journal={Advances in Neural Information Processing Systems},
  volume={37},
  pages={108042--108087},
  year={2024}
}

@inproceedings{somepalli2023diffusion,
  title={Diffusion art or digital forgery? investigating data replication in diffusion models},
  author={Somepalli, Gowthami and Singla, Vasu and Goldblum, Micah and Geiping, Jonas and Goldstein, Tom},
  booktitle={Conference on Computer Vision and Pattern Recognition},
  pages={6048--6058},
  year={2023},
}

@article{webster2023duplication,
  title={On the de-duplication of laion-2b},
  author={Webster, Ryan and Rabin, Julien and Simon, Loic and Jurie, Frederic},
  journal={arXiv preprint arXiv:2303.12733},
  year={2023}
}

@inproceedings{birhane2024dark,
  title={The dark side of dataset scaling: Evaluating racial classification in multimodal models},
  author={Birhane, Abeba and Dehdashtian, Sepehr and Prabhu, Vinay and Boddeti, Vishnu},
  booktitle={ACM Conference on Fairness, Accountability, and Transparency},
  pages={1229--1244},
  year={2024}
}

@inproceedings{hong2024s,
  title={Who's in and who's out? A case study of multimodal CLIP-filtering in DataComp},
  author={Hong, Rachel and Agnew, William and Kohno, Tadayoshi and Morgenstern, Jamie},
  booktitle={ACM Conference on Equity and Access in Algorithms, Mechanisms, and Optimization},
  pages={1--17},
  year={2024}
}

@article{moore2020fiftyone,
  title={FiftyOne},
  author={Moore, B. E. and Corso, J. J.},
  journal={GitHub. Note: https://github.com/voxel51/fiftyone},
  year={2020}
}

@misc{clip_based_nsfw_detector,
  author = {LAION-AI},
  title = {CLIP-based-NSFW-Detector},
  year = {2022},
  publisher = {GitHub},
  journal = {GitHub repository},
  howpublished = {\url{https://github.com/LAION-AI/CLIP-based-NSFW-Detector}}
}

@misc{beaumont-2021-img2dataset,
  author = {Romain Beaumont},
  title = {img2dataset: Easily turn large sets of image urls to an image dataset},
  year = {2021},
  publisher = {GitHub},
  journal = {GitHub repository},
  howpublished = {\url{https://github.com/rom1504/img2dataset}}
}

@article{raffaelc4,
  title={Exploring the limits of transfer learning with a unified text-to-text transformer},
  author={Raffel, Colin and Shazeer, Noam and Roberts, Adam and Lee, Katherine and Narang, Sharan and Matena, Michael and Zhou, Yanqi and Li, Wei and Liu, Peter J},
  journal={Journal of machine learning research},
  volume={21},
  number={140},
  pages={1--67},
  year={2020}
}

@inproceedings{wang2018glue,
  title={GLUE: A multi-task benchmark and analysis platform for natural language understanding},
  author={Wang, Alex and Singh, Amanpreet and Michael, Julian and Hill, Felix and Levy, Omer and Bowman, Samuel R},
  booktitle={EMNLP workshop BlackboxNLP: Analyzing and interpreting neural networks for NLP},
  pages={353--355},
  year={2018}
}

@inproceedings{zhu2015bookcorpus,
  title={Aligning books and movies: Towards story-like visual explanations by watching movies and reading books},
  author={Zhu, Yukun and Kiros, Ryan and Zemel, Rich and Salakhutdinov, Ruslan and Urtasun, Raquel and Torralba, Antonio and Fidler, Sanja},
  booktitle={International Conference on Computer Vision},
  pages={19--27},
  year={2015}
}

@article{haraway1988situatedknowledges,
  author  = {Donna Haraway},
  title   = {Situated Knowledges: The Science Question in Feminism and the Privilege of Partial Perspective},
  journal = {Feminist Studies},
  volume  = {14},
  number  = {3},
  pages   = {575--599},
  year    = {1988},
}

@article{van2016stereotyping,
  title={Stereotyping and bias in the flickr30k dataset},
  author={Van Miltenburg, Emiel},
  journal={arXiv preprint arXiv:1605.06083},
  year={2016}
}

@article{schuhmann2021laion,
  title={Laion-400m: Open dataset of clip-filtered 400 million image-text pairs},
  author={Schuhmann, Christoph and Vencu, Richard and Beaumont, Romain and Kaczmarczyk, Robert and Mullis, Clayton and Katta, Aarush and Coombes, Theo and Jitsev, Jenia and Komatsuzaki, Aran},
  journal={arXiv preprint arXiv:2111.02114},
  year={2021}
}

\end{document}